\documentclass[letterpaper, 10 pt, conference]{ieeeconf}  %

\IEEEoverridecommandlockouts                              %

\usepackage{graphics} %

\usepackage{amsmath}
\usepackage{amsfonts}
\usepackage{graphicx}
\usepackage{subcaption}
\usepackage{caption}
\usepackage{xspace}
\usepackage{multirow}
\usepackage{booktabs}
\usepackage{todonotes}
\usepackage{pifont}
\usepackage{algorithmicx}
\usepackage{algorithm}
\usepackage{wasysym}
\usepackage{placeins}
\usepackage[noend]{algpseudocode}
\usepackage{adjustbox}
\usepackage{threeparttable}
\newcommand{\cmark}{\ding{51}}
\newcommand{\xmark}{\ding{55}}%

\newcommand{\ourmodel}{Spatially-Aware Graph Neural Networks} %
\newcommand{\ourmodelshort}{\textsc{SpAGNN}}
\newcommand{\ourdataset}{\textsc{ATG4D}}
\newcommand{\nuscenes}{\textsc{nuScenes}}

\makeatletter
\DeclareRobustCommand\onedot{\futurelet\@let@token\@onedot}
\def\@onedot{\ifx\@let@token.\else.\null\fi\xspace}

\def\eg{\emph{e.g}\onedot} 
\def\ie{\emph{i.e}\onedot}

\makeatother

\title{\LARGE \bf
\ourmodelshort{}: Spatially-Aware Graph Neural Networks \\ for Relational Behavior Forecasting from Sensor Data
}

\author{
\textbf{Sergio Casas$^{1,2}$, Cole Gulino$^{1}$, Renjie Liao$^{1,2}$, Raquel Urtasun$^{1,2}$} \\
Uber Advanced Technologies Group$^{1}$, University of Toronto$^{2}$ \\
\texttt {\{sergio.casas, cgulino, rjliao, urtasun\}@uber.com}
}

\begin{document}

\maketitle
\thispagestyle{empty}
\pagestyle{empty}

\begin{abstract}
In this paper, we tackle the problem of relational behavior forecasting from sensor data.
Towards this goal, we propose a novel spatially-aware graph neural network (\ourmodelshort{}) that models the interactions between agents in the scene. 
Specifically, we  exploit a convolutional neural network to detect the actors and compute their initial states.
A graph neural network then iteratively updates the actor states via a message passing process.
Inspired by Gaussian belief propagation, we design the messages to be  spatially-transformed parameters of the output distributions from neighboring agents. 
Our model is fully differentiable, thus enabling end-to-end training.
Importantly, our probabilistic predictions can model uncertainty at the trajectory level.
We demonstrate the effectiveness of our approach by achieving significant improvements over the state-of-the-art on two real-world self-driving datasets: \ourdataset{} and \nuscenes{}.

\end{abstract}

\section{Introduction}

Self-driving is one of the most exciting challenges of contemporary artificial intelligence because of its potential to revolutionize transportation. 
While there has been incredible progress in machine learning and robotics in the past years, many challenges still remain to achieve full autonomy.
One of the critical challenges is that self-driving vehicles will need to share the space with human drivers, who can perform a very diverse set of maneuvers 
including compromising behaviors. Particularly, these maneuvers and behaviors are highly determined by the interactions with neighboring drivers \cite{wilde1976social, mcnabb2017ll, connolly1993some}.

Understanding human intention is a very difficult task. In order to predict the other drivers' future behavior it is necessary to perceive their past motion, 
analyze the  interplay with other agents and process the information available from the scene such as the lane-graph. 
Therefore, for autonomous vehicles to be able to coexist with human drivers in the roads, they need to be able to emulate human behaviors \cite{casas2018intentnet, bansal2018chauffeurnet, zeng2019end}. 
For this reason, the problem of detection and long-term future behavior forecasting of 
other vehicles in realistic environments is at the core of safe motion planning. %

In the past few years many deep learning approaches have been developed to detect objects  from LiDAR point clouds \cite{li2016vehicle, yang2018pixor, engelcke2017vote3deep, yang2018hdnet}. 
They mostly differ in the input representation and the architecture.  
Voxels \cite{engelcke2017vote3deep, luo2018fast}, bird's eye view  \cite{yang2018pixor, yang2019std, liang2019multi} or range view \cite{li2016vehicle, chen2017multi, meyer2019lasernet} representations are typically employed.
Recent work \cite{djuric2018motion, bansal2018chauffeurnet, cui2018multimodal} showed how to utilize convolutional neural networks (CNNs) to produce future trajectories of actors given 3D object detections as input.
However, this approach cannot recover from mistakes at the detection stage.
A recent seminal work \cite{luo2018fast} proposed to jointly perform object detection and motion forecasting within the same neural network. 
This was further extended to also reason about actor intentions \cite{casas2018intentnet}. 
However, all these approaches ignore the social interactions between the agents. %

\begin{figure}
    \centering
    \resizebox{\columnwidth}{!}{%
    \includegraphics[width=\textwidth, trim={4cm 2cm 4cm 2cm}, clip]{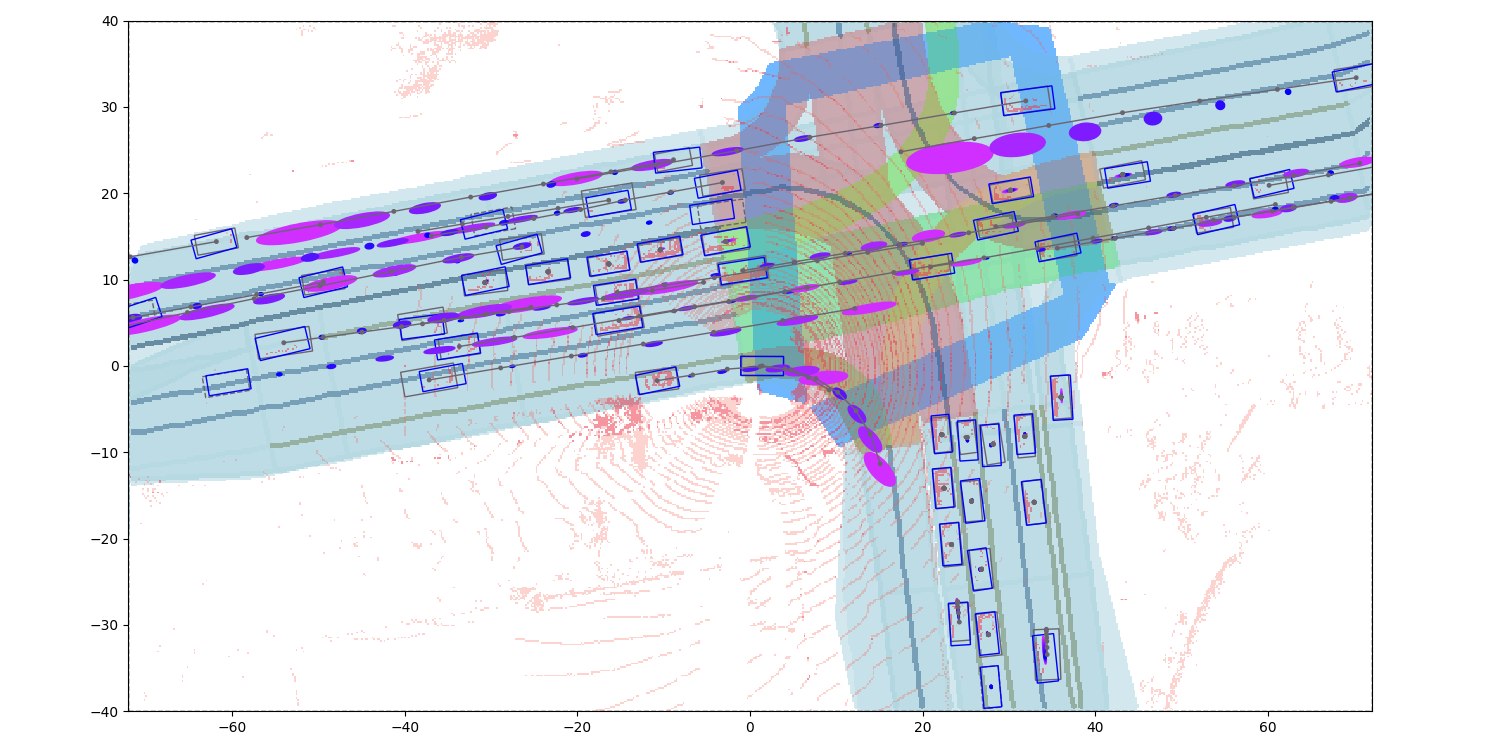}
    \vspace{-0.7cm}
    }
    \caption{\ourmodelshort{} takes as input LiDAR point clouds (red dots) and HD Maps (road in light blue, traffic lights in green/red, etc.) 
             and outputs detections (blue bounding boxes) as well as a \textbf{socially coherent probabilistic motion forecasting} (blue to pink indicates time).
             Ground-truth boxes and future waypoints are displayed in gray.}
    \label{fig:qualitative}
    \vspace{-0.7cm}  
\end{figure}

Interactions are very common in real-world driving.
Illustrative examples are the negotiation between drivers that take place in 4-way stop intersections, yielding situations like unprotected left turns, and even the very simple car-following behavior.
Modeling these interactions will help significantly to reduce the uncertainty in predicting the future behavior. 
Towards this goal, we propose an efficient probabilistic model that leverages recent advances in graph neural networks (GNNs) \cite{scarselli2008graph} to capture the interactions between vehicles. 
Our model is fully differentiable, enabling the joint optimization of detection and behavior forecasting tasks, which mitigates the propagation of early errors.

We showcase the power of our approach 
by showing significant improvements over the  state-of-the-art across all
detection, motion forecasting, and interaction metrics in both \ourdataset{} \cite{yang2018pixor}, a large-scale dataset with over a million frames that we collected, as well as the newly released \nuscenes{}  \cite{caesar2019nuscenes} dataset.
In the remainder of the paper, we first give an overview of the related work, then present our model, and finally discuss our 
experimental setup, exhibiting both quantitative and qualitative results.

\section{Related Work}

In this section, we first review recent advances in object detection from point clouds. 
We then  discuss motion estimation approaches as well as methods for agent interaction modeling.  
Finally, we review  joint perception and behavior forecasting methods.

\paragraph{Object Detection from Point Clouds} 
A popular approach is to use 3D convolutional networks that operate over voxel grids  \cite{engelcke2017vote3deep, li20173d, luo2018fast}. 
However, the sparsity of point clouds makes the computation redundant. 
Front-view representations \cite{li2016vehicle, chen2017multi, meyer2019lasernet} exploiting  range information of a LiDAR sensor have also been explored with success, although they lose the original metric space and have to handle large variations in object size caused by projection.  %
Another option is to handle point clouds directly (without voxelization) \cite{qi2017pointnet, qi20173d, qi2017pointnet++, wang2018deep}. 
Unfortunately, all the above methods suffer from either limited performance or heavy computation \cite{simon1803complex}. 
Recently, bird's-eye-view  detectors that exploit 2D convolutions over the ground-plane \cite{yang2018pixor, yang2018hdnet, yang2019std, liang2019multi}  have shown superior performance in terms of speed and accuracy. 

\paragraph{Motion Forecasting} 
DESIRE \cite{lee2017desire} proposed a recurrent variational auto-encoder to generate trajectories from ground-truth past trajectories and images. 
R2P2 \cite{rhinehart2018r2p2} proposed a flow-based generative model to learn object dynamics.
However, both DESIRE and R2P2 are not ideal for time-critical applications due to their expensive sampling needed to cover all possible outcomes.
SIMP \cite{hu2018probabilistic} parametrizes the output space as insertion areas where vehicle could go, predicting an estimated time of arrival and a spatial offset. 
\cite{djuric2018motion}, \cite{bansal2018chauffeurnet} and \cite{cui2018multimodal} create bird's eye view rasters from the lane graph of the scene and 
perception results produced by a separate system to predict future trajectories of road users. 
Unfortunately, the actor-centric rasterization employed poses a challenge for real-time applications. Another limitation of these methods is that  perception and motion forecasting modules  are learned separately. 

\paragraph{Interaction Modeling} 
\cite{ma2017forecasting} proposed to couple game theory and deep learning to model the social aspect of pedestrian behavior. 
Several methods  have exploited \cite{alahi2016social, santoro2017simple, sun2018actor, sun2019relational, deo2018convolutional,  rhinehartprecog, gupta2018social, sadeghian2019sophie} a variety of social pooling layers to include relational reasoning in convolutional and recurrent neural networks. 
Graph neural networks (GNNs) \cite{li2017situation, schlichtkrull2018modeling, kipf2018neural} have recently been shown to be very effective.
NRI \cite{kipf2018neural}  models the interplay of components by using GNNs to explicitly infer interactions while simultaneously learning the dynamics. %
CAR-Net \cite{sadeghian2018car} models agent-scene interactions by coupling two specialized attention mechanisms.
However, all the aforementioned methods still assume perfect perception when facing the behavior forecasting task. 

\paragraph{Joint Perception and Behavior Forecasting} 
FaF \cite{luo2018fast} unified object detection and short term motion forecasting from LiDAR. 
IntentNet \cite{casas2018intentnet} modified FaF's architecture, replacing 3D by 2D convolutions and adding the prediction of high-level intentions for each agent by exploiting rich semantic information from HD maps. 
This was further extended to also predict a cost map for ego-vehicle motion planning \cite{zeng2019end}. 
However, although these models perform future behavior prediction of vehicles in urban scenes, they do not model multi-agent interactions explicitly. 
Here we show how to leverage the success of joint detection and prediction while reasoning about interactions between agents.

\begin{figure*}
        \centering
        \includegraphics[page=1, trim={0cm 0.08cm 0cm 0.04cm}, clip, width=0.8\textwidth]{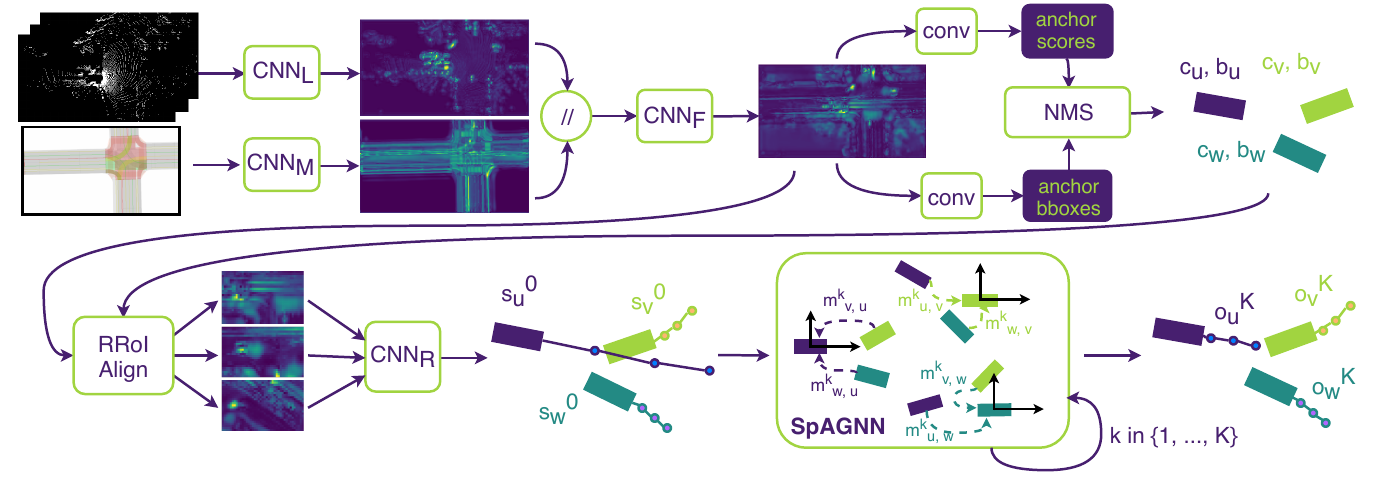}
        \vspace{-0.15cm}
        \caption{
        \textbf{Inference diagram of \ourmodelshort{}.}
        It consists of two stages: object detection (Section \ref{sec:detection}) in the top row and relational behavior forecasting (Section \ref{sec:pred}) in the bottom row. 
        }	
        \label{fig:architecture}
    \vspace{-0.4cm}    
    \end{figure*}

\section{Object Detection from LiDAR and HD Maps} \label{sec:detection}

In this section, we first discuss our input parametrization to exploit 3D LiDAR  and HD maps. We then explain the first stage of our model, 
\ie, the backbone network for object detection (top row in Fig.\ref{fig:architecture}).

We voxelize the 3D LiDAR point cloud similarly to \cite{yang2018pixor}, 
with the difference that we leverage ground height information available in our HD maps to obtain our voxelized LiDAR \cite{yang2018hdnet}. Compared to a sensor-relative height, this reduces the variance in the Z coordinates of vehicles since these always lie on the ground, allowing our model to learn height priors. In order to exploit motion information, we follow \cite{luo2018fast} and  
leverage multiple LiDAR sweeps by projecting the past sweeps to the coordinate frame of the current sweep, taking into account the ego-motion. 
Following \cite{casas2018intentnet}, we stack  the height and time dimensions into the channel dimension of our tensor in order to exploit 2D convolutions.
This provides us with a Bird's Eye View (BEV) 3D occupancy tensor of dimensions $\left(\frac{L}{\Delta L}, \frac{W}{\Delta W}, \frac{H}{\Delta H} \cdot T\right)$,
where $\text{L=140}$, $\text{W=80}$ and $\text{H=5}$ meters are the longitudinal, transversal and normal physical dimensions of the scene we employ in the \ourdataset{} dataset. We reduce the region of interest to 50 by 50 meters in \nuscenes{} due to the limited range of its 32-beams LiDAR sensor as well as the annotations. $\Delta \text{L=}\Delta \text{W=}\Delta \text{H=0.2}$ meters/pixel are the voxel sizes in the corresponding directions and $\text{T=10}$ is the number of LiDAR sweeps we employ in both datasets.

Following \cite{casas2018intentnet}, our input raster map contains information regarding roads, lanes, intersections, crossings, traffic signs and traffic lights\footnote{\scriptsize We use an image-based CNN  to estimate the state of the traffic light.}.
\normalsize
In such a representation, different semantics are encoded in separate channels to ease the learning of the CNN and avoid predefining orderings in the raster. 
For instance, yellow markers denoting the barrier between opposing traffic are rasterized in a different channel than white markers. 
In total, this representation consists of 17 binary channels.

We build our object detection network on top of PIXOR \cite{yang2018pixor}, which is a lightweight state-of-the-art object detector. 
In particular, we extend the single branch network of PIXOR to a two-stream network such that one stream processes LiDAR point clouds 
and the other processes HD maps, respectively referred as $\operatorname{CNN_L}$ and $\operatorname{CNN_M}$ in Fig. \ref{fig:architecture}.
We modify PIXOR's backbone by first reducing the number of layers in the first 4 residual blocks from (3, 6, 6, 4) to (2, 2, 3, 6) in order to save computation.
LiDAR point clouds are fed to this condensed backbone. To process the high-definition map, we replicate this backbone but halve the number of filters at each layer for efficiency purposes.
After extracting features from the LiDAR and HD map streams, we concatenate them along the channel dimension. 
The concatenated features are then fused by a header convolutional network ($\operatorname{CNN_F}$).
Two convolutional layers are then used to output a confidence score and a bounding box for each anchor location, 
which are further reduced to the final set of candidates by applying non-maximum suppression (NMS).
These modifications allow us to create a high performing detector that is also very fast.

\section{Relational Behavior Forecasting} \label{sec:pred}

The second stage of our model provides a  probabilistic formulation for predicting the future states of  detected vehicles by exploiting the interactions between different actors. 
We denote the  $i$-th actor state at time $t$ as $s_{i,t} = (\mathbf{x}_{i,t}, \theta_{i,t})$. 
The state includes a future trajectory composed of 2D waypoints $\mathbf{x}_{i,t}$ and heading angles $\theta_{i,t}$. 
Let $\Omega$ be the scene input  composed of LiDAR and HD map. The number of detected actors in a scene is denoted as $N$ and the future time steps to be predicted as $T$.
Note that the number of actors $N$ varies from  scene to scene and our relational model is general and works for any cardinality. %
As the number of vehicles in the scene is  not large (typically less than a hundred), we use a fully connected directed graph to let the model
figure out the importance of the interplay for each pair of actors in a bidirectional fashion. 
Bidirectionality is important as  the relationships are asymetric, e.g., vehicle following the vehicle in front.  

To design our probabilistic relational behavior forecasting approach, we take inspiration from the Gaussian Markov random field (Gaussian MRF) and design a novel type of graph neural network for this task. 
In the following, we first describe the Gaussian MRF and its canonical inference algorithm Gaussian belief propagation (GaBP). 
We then briefly introduce graph neural networks (GNNs) and dive deep into our \ourmodel{} (\ourmodelshort{}). %

\subsection{Gaussian MRFs and Gaussian Belief Propagation}

We now introduce the Gaussian MRF  and its inference algorithm Gaussian belief propagation in our problem context. 
Conditioned on the observed input  and detection output, we assume the future states can be predicted independently for different future time steps.
How to explore temporal dependency is left as future work.
Therefore, from now on, we drop the subscript of time for simplicity.
In a Gaussian MRF, the joint probability is assumed to be a multivariate Gaussian distribution, \ie, $p(s_{1}, \cdots, s_{N} | \Omega) \propto \exp( \mathbf{s}^{\top} \mathbf{A} \mathbf{s} +  \mathbf{b}^{\top} \mathbf{s})$ where $\mathbf{s}$ is the concatenation of all $s_{i}$, and  
$\mathbf{A}$ and $\mathbf{b}$ are the model parameters.
Based on the interaction graph, we can decompose the joint probability as follows,
\begin{align}\label{eq:gauss_mrf}
	p(s_{1}, \cdots, s_{N} | \Omega) \propto \prod_{i} \phi_{i}(s_{i}, \Omega) \prod_{ij} \psi_{ij}(s_{i}, s_{j}, \Omega) 	
\end{align}
where the unary and pairwise potentials are,
{%
\begin{align}\label{eq:potentials}	
	\phi_{i}(s_{i}, \Omega) & = \exp(-\frac{1}{2} s_{i}^{\top} A_{ii} s_{i} + b_{i}^{\top} s_{i}) \nonumber \\
	\psi_{ij}(s_{i}, s_{j}, \Omega) & = \exp(-\frac{1}{2} s_{i}^{\top} A_{ij} s_{j}). 	
\end{align}
}
\vspace{-0.1cm}
\normalsize
Note that $A_{ii}$, $b_{i}$ and $A_{ij}$ depend on the input $\Omega$. 
Their specific functional forms can be designed  according to the application.
It is straightforward to show that the unary potentials follow a Gaussian distribution, \ie, $\phi_{i}(s_{i}, \Omega) \propto \mathcal{N}(s_{i} \vert A_{ii}^{-1} b_{i}, A_{ii}^{-1})$.

To compute the marginal distribution $p(s_{i} | \Omega)$, Gaussian belief propagation (GaBP)~\cite{weiss2000correctness} is often adopted for exact inference.
In particular, denoting the mean and precision (inverse covariance) matrix of the message from node $i$ to node $j$ as $\mu_{ij}$ and $P_{ij}$, one can  derive
the following iterative update equations based on the belief propagation algorithm and Gaussian integral:
{\footnotesize
\begin{align}\label{eq:gauss_bp}
	P_{ij} & = - A_{ij}^{-1} ( A_{ii} + \sum\nolimits_{k \in \mathbf{N}(i) \backslash j} P_{ki} ) A_{ij}^{-1} \nonumber \\
	\mu_{ij} & = - P_{ij}^{-1} A_{ij} ( A_{ii} + \sum_{k \in \mathbf{N}(i) \backslash j} P_{ki} )^{-1} ( b_{i} + \sum_{k \in \mathbf{N}(i) \backslash j} P_{ki} \mu_{ki} ),
\end{align}
}
\normalsize
where $\mathbf{N}(i)$ is the neighborhood of node $i$ and $\mathbf{N}(i) \backslash j$ is the same set without node $j$.
Once the message passing converges, one can compute the exact marginal mean and precision as below,
{
\begin{align}\label{eq:gauss_mrf_gt_marginal}
	P_{i} & = A_{ii} + \sum\nolimits_{k \in \mathbf{N}(i)} P_{ki} \nonumber \\
	\mu_{i} & = P_{i}^{-1} (b_{i} + \sum\nolimits_{k \in \mathbf{N}(i)} P_{ki} \mu_{ki} ),
\end{align}
}
\vspace{-0.1cm}
\normalsize
where $p(s_{i} | \Omega) = \mathcal{N}(s_{i} \vert \mu_{i}, P_{i}^{-1})$.

\pagebreak

\subsection{\ourmodel{} (\ourmodelshort{})}

Although the Gaussian MRF is a powerful model, it has important limitations in our scenario.
First, some of our states  (\ie, the heading angle) can not be represented as a Gaussian random variable due to its bounded support between $-\pi$ and $+\pi$. 
Second, for non-Gaussian data, the integral in the belief propagation update is generally intractable.
However, the Gaussian MRF and the GaBP algorithm give us great inspiration in designing our approach. 
In the following, we first briefly review graph neural networks (GNNs) and then describe our novel formulation.

Graph Neural Networks (GNNs) \cite{scarselli2008graph} are powerful models for processing graph-structured data because (1) the model size does not depend on the input graph size (interaction graphs in our case have varying sizes) and (2) they have high capacity to learn good representations both at a node and graph level.
Given an input graph and node states, a GNN unrolls a finite-step message passing algorithm over the graph to update node states.
In particular, for each edge, one first computes a message  
in parallel via a shared message function which is a neural network taking the state of the two terminal nodes as input. 
Then, each node aggregates the incoming messages from its local neighborhood using an aggregation operator, \eg, summation.
Finally, each node updates its own state based on its previous state and the aggregated message using another neural network.
This message passing is repeated for a finite number of times for practical reasons.

In our context, we consider each actor to be a node in the interaction graph.
If we view the node state as mean and precision matrix of the marginal Gaussian as in Gaussian MRFs, what GaBP does is very similar to what a GNN does.
Specifically, computing and updating messages as in Eq. (\ref{eq:gauss_bp}, \ref{eq:gauss_mrf_gt_marginal}) can be regarded as particular instantiations of graph neural networks.
Therefore, one can generalize the message passing of GaBP using a GNN based on the universal approximation capacity of neural networks.
Note that not all instantiations of GNN will guarantee the convergence to the true marginal as GaBP does in the Gaussian MRFs. 
Nonetheless, GNNs can be trained using back-propagation and can effectively handle non-Gaussian data thanks to their high capacity. 
Motivated by the similarity between GaBP and GNN, we design \ourmodelshort{} as described below.

\paragraph{Node State}

The node state of our \ourmodelshort{} consists of two parts that will be updated iteratively: a hidden state an an output state.
For the $v$-th node, we construct the initial hidden state $h_{v}^{(0)}$ by extracting the region of interest (RoI) feature map from the detection backbone network 
for the $v$-th detection.
In particular, we first apply the recently proposed Rotated RoI Align \cite{ma2018arbitrary}, an improved variant of the previously proposed RoI pooling \cite{girshick2015fast} 
and RoI align \cite{he2017mask} that extracts fixed size spatial feature maps for bounding boxes with arbitrary shapes and rotations.
We then apply a 4-layer down-sampling convolutional network followed by max pooling to reduce the 2D feature map to a 1D feature vector per actor ($\operatorname{CNN_R}$ in Fig. \ref{fig:architecture}).

Inspired by GaBP, the output state $o_{v}^{(k)}$ at each message passing step $k$ consists of statistics of the marginal distribution.
Specifically, we assume the marginal of each waypoint and angle follow a Gaussian and Von Mises distributions respectively, \ie, $p(\mathbf{x}_{v}^{(k)} | \Omega) = \mathcal{N}(\mathbf{x}_{v}^{(k)} \vert \boldsymbol{\mu}_{v}^{(k)}, \boldsymbol{\Sigma}_{v}^{(k)})$, $p(\theta_{v}^{(k)} | \Omega) = \mathcal{V}(\theta_{v}^{(k)} \vert \eta_{v}^{(k)}, \kappa_{v}^{(k)})$, where
$\mathbf{x}_{v}^{(k)} = [ {x_{v}^{(k)}}, {y_{v}^{(k)}} ]^{\top}$, $\boldsymbol{\mu}_{v}^{(k)} = [ \mu_{x_{v}}^{(k)}, \mu_{y_{v}}^{(k)} ]^{\top}$,
{\small
\begin{align}
\boldsymbol{\Sigma}_{v}^{(k)} = \left(\begin{array}{cc}{{\sigma_{x_{v}}^{(k)}}^{2}} & {\rho_{v}^{(k)} \sigma_{x_{v}}^{(k)} \sigma_{y_{v}}^{(k)}} \\ {\rho_{v} \sigma_{x_{v}}^{(k)} \sigma_{y_{v}}^{(k)}} & {{\sigma_{y_{v}}^{(k)}}^{2}}\end{array}\right).
\end{align}
}
Therefore, the output state predicted by our model $o_{v}^{(k)}$ is the concatenation of the parameters of both distributions $\mu_{x_{v}}^{(k)}$, $\mu_{y_{v}}^{(k)}$, $\rho_{v}^{(k)}$, $\sigma_{x_{v}}^{(k)}$, $\sigma_{y_{v}}^{(k)}$, $\eta_{v}^{(k)}$ and $\kappa_{v}^{(k)}$.
The goal is to gradually improve the output states in the GNN as the message passing algorithm goes on.
Note that we evaluate the likelihood using the local coordinate system centered at each actor and oriented in a way that the x axis is aligned with the heading direction, as shown in Fig. \ref{fig:architecture}.
This makes the learning task easier compared to using a global anchor coordinate system like in \cite{luo2018fast,casas2018intentnet}; as shown in \cite{chou2018arxiv}.
To initialize the output state $o_{v}^{(0)}$, we use a 2-layer MLP which takes the max-pooled RoI features $h_{v}^{(0)}$ as input and directly predicts the output state, independently per actor.

\paragraph{Message passing}
The node states are iteratively updated by a message passing process.
For any directed edge $(u, v)$, at propagation step $k$, we compute the message $m_{u \rightarrow v}^{(k)}$ as 
{\small
\begin{align}\label{eq:gnn_msg}
m_{u \rightarrow v}^{(k)} = \mathcal{E}^{(k)}\left(h_{u}^{k-1}, h_{v}^{k-1}, \mathcal{T}_{u,v} (o_{u}^{(k-1)}), o_{v}^{(k-1)}, b_{u}, b_{v} \right)
\end{align}
}
where $\mathcal{E}^{(k)}$ is a 3-layer MLP and $\mathcal{T}_{u,v}$ is the transformation from the coordinate system of detected box $b_{u}$ to the one of $b_{v}$.
Note that we rotate the state $o_{u}^{(k)}$ for each neighbor of node $v$ such that they are relative to the local coordinate system of $v$.
By doing so, the model is aware of the spatial relationship between two actors, which eases the learning taking into account that is extremely hard to extract such information from local, 
RoI pooled features. We show the advantages of projecting the output state of node $u$ to the local coordinate system of node $v$ when computing the message $m_{u \rightarrow v}^{(k)}$ in 
the ablation study in Table \ref{table:ablation}. After computing the messages on all edges, we aggregate the messages going to node $v$ as follows,
{\small
\begin{align}\label{eq:gnn_msg_agg}
a_{v}^{(k)}=\mathcal{A}^{(k)}\left(\left\{m_{u \rightarrow v}^{(k)} : u \in \mathbf{N}(v)\right\}\right),
\end{align}
}
We use an ordering-invariant, feature-wise $\operatorname{max}$ operator along the neighborhood dimension as $\mathcal{A}^{(k)}$ function.

\paragraph{State Update}
Once we compute the aggregated message $a_{v}^{(k)}$, we can update the node state  
{\small
\begin{align}\label{eq:gnn_update}
h_{v}^{(k)}=\mathcal{U}^{(k)}\left(h_{v}^{(k-1)}, a_{v}^{(k)}\right) \quad
o_{v}^{(k)}=\mathcal{O}^{(k)}\left(h_{v}^{(k)}\right)
\end{align}
}
where $\mathcal{U}^{(k)}$ is a GRU cell and $\mathcal{O}^{(k)}$ is a 2-layer MLP.

The above message passing process is unrolled for $K$ steps,  where $K$ is a hyperparameter. The final prediction of the model is $O^{K} = \{o_{v}^{(K)}\}$. 
Note that the design of the message passing algorithm in Eq. (\ref{eq:gnn_msg}, \ref{eq:gnn_msg_agg}, \ref{eq:gnn_update}) 
can be regarded as generalization of the one in Eq. (\ref{eq:gauss_bp}, \ref{eq:gauss_mrf_gt_marginal}) due to the universal approximation capacity of neural networks.

\pagebreak

\section{End-to-end Learning}

Our full model (including detection and relational prediction) is trained jointly end-to-end through back-propagation. 
In particular, we minimize a multi-task objective containing a binary cross entropy loss for the classification branch of the detection network (background vs vehicle), 
a regression loss to fit the detection bounding boxes and a negative log likelihood term for the probabilistic trajectory prediction.
We apply hard negative mining to our classification loss: we select all positive examples from the ground-truth and 3 times as many negative examples from the rest of anchors. 
Regarding our box fitting, we apply a smooth L1 loss \cite{girshick2015fast} to each of the 5 parameters $(x, y, w, h, \text{sin}(\theta), \text{cos}(\theta))$ of the bounding boxes anchored to a positive example. 
The negative log-likelihood (NLL) is as follows:
{\small
\begin{align*}
\mathcal{L}_{nll} &= \sum_{i=1}^N  \sum_{t=1}^T \frac{1}{2} \log|\boldsymbol{\Sigma}_{i,t}| + \frac{1}{2}(\mathbf{x}_{i,t}-\boldsymbol{\mu}_{i,t})^{\mathrm{T}} \boldsymbol{\Sigma}_{i,t}^{-1}(\mathbf{x}_{i,t}-\boldsymbol{\mu}_{i,t}) \\
&- \kappa_{i,t} \cos (\theta_{i,t}-\eta_{i,t}) + \log \left( 2 \pi I_{0}(\kappa_{i,t}) \right)
\end{align*}
}
where the first line corresponds to the NLL of a 2D gaussian distribution and the second line to the NLL of a Von Mises distribution, $I_{0}$ being the modified Bessel function of order 0.
For the GNN message passing, we use back-propagation through time to pass the gradient to the detection backbone network.

\section{Experimental evaluation}

In this section, we first explain the datasets and metrics that we use for evaluation. Next, we compare our model against  state-of-the-art detection and motion forecasting algorithms. We then perform an ablation study to understand what contributes the most to the performance gain of our \ourmodelshort{}. 
Finally, we show some qualitative results. We defer the implementation details of our method and the baselines to the appendix (\ref{implementation_details} and \ref{baselines_details}).

\paragraph{Datasets} We report results on two datasets: \ourdataset{} and the recently released \nuscenes{} \cite{caesar2019nuscenes} dataset. This allow us to test the effectiveness of our approach in two vehicle platforms with different LiDAR sensors driving in different cities. 

We collected the \ourdataset{} dataset  by driving a fleet of self-driving cars over several cities in North America with a 64-beam, roof-mounted LiDAR.  
It contains over 1 million frames collected from 5,500 different scenarios, which are sequences of 250 frames captured at 10 Hz. 
Our labels are very precise tracks of 3D bounding boxes with a maximum distance of a 100 meters.
The \nuscenes{} dataset consists of 1,000 snippets of 20 seconds each, collected in Boston or Singapore. Their 32-beam LiDAR captures a sparser point cloud than the one in \ourdataset. Despite their high sensor capture frequency of 20Hz, only keyframes at 2Hz are employed for  annotation, thus limiting the number of frames available as supervision by an order of magnitude. 
They also provide HD maps of the two cities. 

\paragraph{Metrics}
We evaluate  detection  using the standard precision-recall (PR) curves and its associated mean average precision (mAP) metric. 
Following previous works, we ignore vehicles without any LiDAR point in the current sweep during evaluation.
To showcase the abilities to capture social interaction, 
we use the cumulative collision rate over time, defined as the percentage of predicted trajectories overlapping in space-time. 
A model that identifies interactions properly should achieve a lower collision rate since our dataset does not contain any colliding examples.
To benchmark the  forecasting ability we use the centroid L2 error as well as the absolute heading error at several future horizons. 
It is worth noting that the prediction metrics depend upon the operating point that we choose for the detector (its confidence score threshold) since these metrics can only be computed on true positive detections, i.e. those that get matched to ground-truth labels. 
To make it fair for models with different detection PR curves, 
the motion forecasting and social compliance metrics are computed at a common recall point.

\begin{figure}[t]
		\centering
		\begin{subfigure}{.5\columnwidth}
		  \centering
		  \includegraphics[width=.99\linewidth]{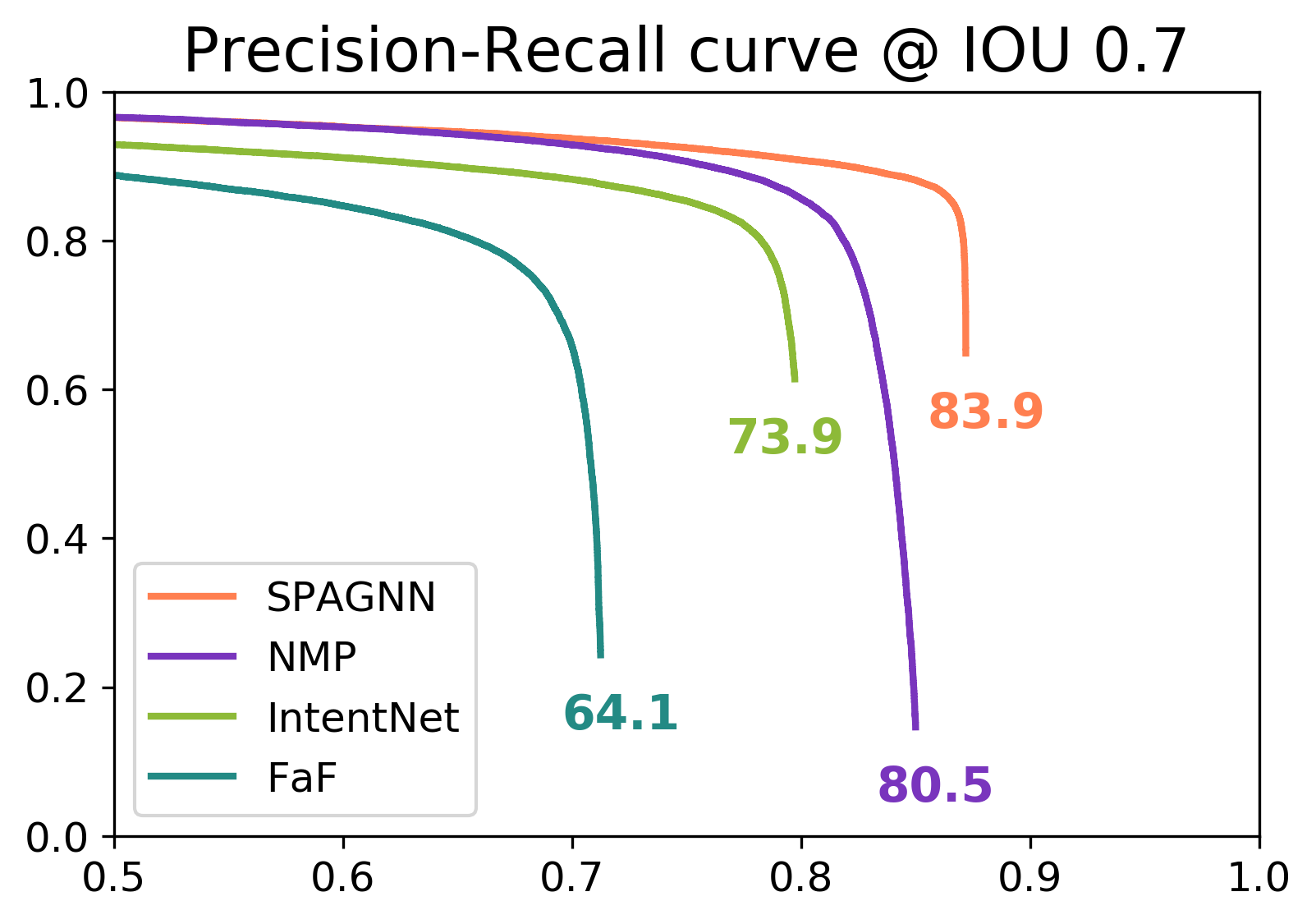}
		\end{subfigure}%
		\begin{subfigure}{.5\columnwidth}
		  \centering
		  \includegraphics[width=.99\linewidth]{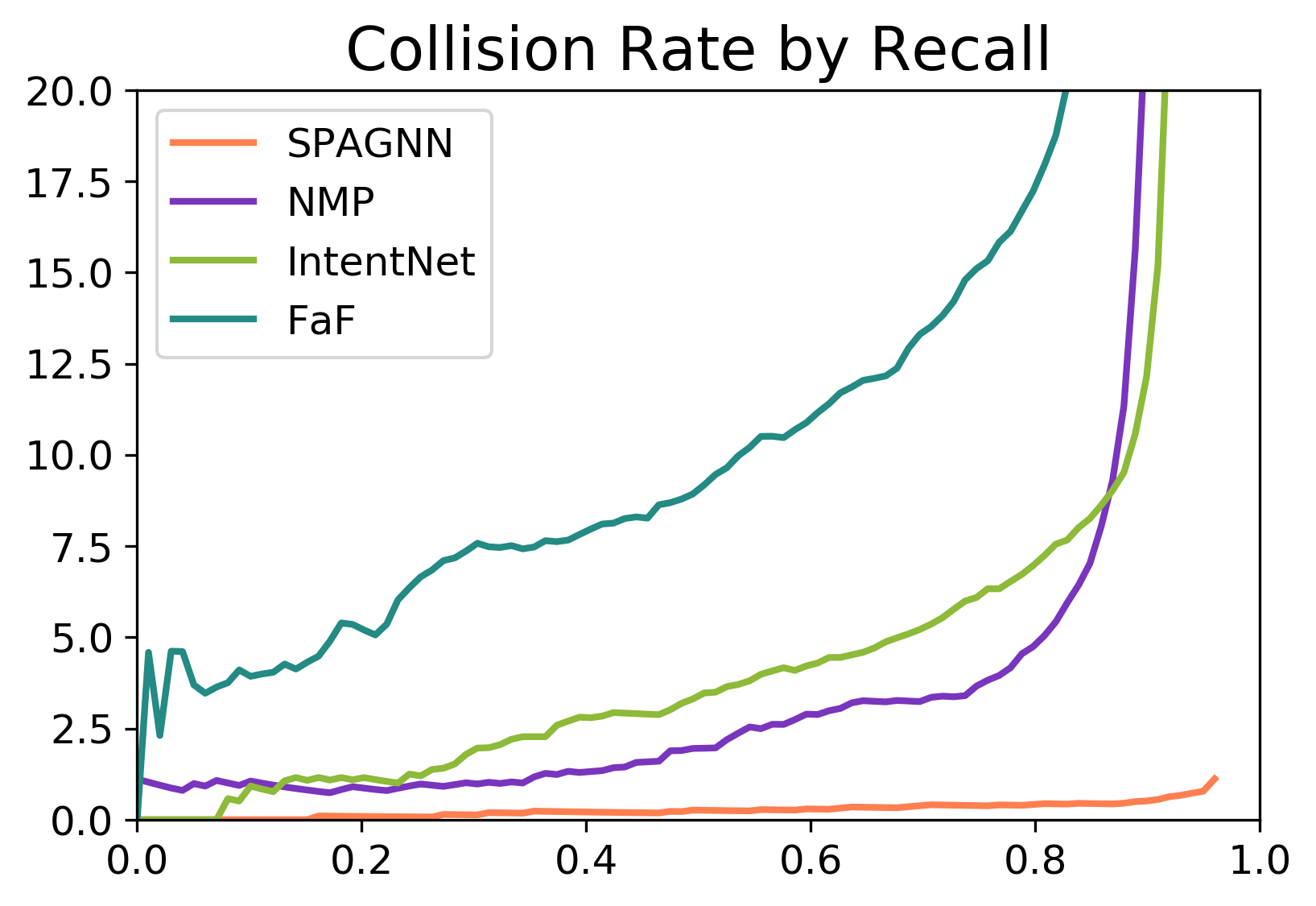}
		\end{subfigure}
		\vspace{-0.1cm}
			\caption{\textbf{[\ourdataset{}] Joint Detection and Prediction}
			}
		\label{fig:main_fig}
	\end{figure}

\begin{table}[t]
	\centering
	\resizebox{\columnwidth}{!}{%
		\small
		\begin{threeparttable}
			\begin{tabular}{c | c @{\hspace{1.0\tabcolsep}} c | c @{\hspace{1.10\tabcolsep}} c @{\hspace{1.10\tabcolsep}} c | c @{\hspace{1.10\tabcolsep}}c@{\hspace{1.10\tabcolsep}}c}
			\toprule
			Model  	                            &  \multicolumn{2}{c}{Col. (\permil)}  & \multicolumn{3}{c}{L2 x,y (cm)} & \multicolumn{3}{c}{Heading err (deg)} 	\\
			{}  	                            & 0-1s  & 0-3s 						&  0s    & 1s 	 &  3s	 	& 0s    & 1s 	&  3s	\\
			\midrule
			D+T+S-LSTM \cite{alahi2016social} 	& 1.43  & 16.31 	 				&  22	 & 147   & 607 		& 4.06  & 5.14	& 8.07		\\ 
			D+T+CSP \cite{deo2018convolutional}	& 1.64  & 20.78 	 				&  22    & 95	 & 282 		& 4.06  & 4.70  & 6.20		\\ 
			D+T+CAR-Net \cite{sadeghian2018car} & 0.28  & 12.30  					&  22    & 46    & 149 		& 4.06  & 4.87  & 6.14 		\\ 
			\midrule
			FaF \cite{luo2018fast}  			& 1.12  & 17.41 	 				&  30    & 54    & 183 		& 4.71  & 4.98	& 6.43		\\ 
			IntentNet \cite{casas2018intentnet} & 0.28  & 7.03 	 					&  26    & 45	 & 146 		& 4.21  & 4.40  & 5.64		\\ 
			NMP \cite{zeng2019end}   			& 0.05  & 3.06   					&  23    & 36  	 & 114 		& 4.10  & 4.24  & 5.09 		\\ 
			\midrule
			E2E S-LSTM \cite{alahi2016social}	& 0.06  & 1.14 	 					&  22	  & 36   & 106 		& 4.97  & 4.85	& 5.61		\\ 
			E2E CSP \cite{deo2018convolutional}	& 0.06  & 4.47 	 					&  23     & 38	 & 114 		& 4.82  & 5.04  & 5.84		\\ 
			E2E CAR-Net \cite{sadeghian2018car} & 0.07  & 1.15  					&  22     & 35   & 105 		& 4.44  & 4.41  & 5.12 		\\ 
			\midrule
			\ourmodelshort{} (Ours)    				& \textbf{0.03} & \textbf{0.42}  &  \textbf{22}    & \textbf{33} 	 &\textbf{ 96} 	  &\textbf{ 3.92}  & \textbf{3.89}	&\textbf{ 4.55 }		\\ 
		
			\bottomrule
			\end{tabular}
			\begin{tablenotes}
				\scriptsize
				\item Legend: D=Detector(PIXOR), T=Tracker(UKF+Hungarian), S-LSTM=SocialLSTM, CSP=Convolutional Social Pooling, E2E=End-to-End, Col=Collision rate
			\end{tablenotes}
		\end{threeparttable}
	}
	\vspace{-0.1cm}
	\caption{\textbf{[\ourdataset] Social interaction and motion forecasting metrics} at 80\% detection recall}
	\label{table:main_table}
	\vspace{-0.3cm}
\end{table}

\begin{table}[t]
	\centering
	\resizebox{\columnwidth}{!}{%
		\small
		\begin{threeparttable}
			\begin{tabular}{c | c @{\hspace{1.0\tabcolsep}} c | c @{\hspace{1.10\tabcolsep}} c @{\hspace{1.10\tabcolsep}} c | c @{\hspace{1.10\tabcolsep}}c@{\hspace{1.10\tabcolsep}}c}
			\toprule
			Model  	                            &  \multicolumn{2}{c}{Col. (\permil)}  & \multicolumn{3}{c}{L2 x,y (cm)} & \multicolumn{3}{c}{Heading err (deg)} 	\\
			{}  	                            & 0-1s  & 0-3s 						&  0s    & 1s 	 &  3s	 	& 0s    & 1s 	&  3s	\\
			\midrule
			E2E S-LSTM \cite{alahi2016social}	& 0.84  & 9.64 	 					&  24	  & 71   & 185 		& 3.08  & 3.59	& 4.63		\\ 
			E2E CSP \cite{deo2018convolutional}	& 0.41  & 5.77 	 					&  24     & 70	 & 174 		& 3.14  & 3.51  & 4.64		\\ 
			E2E CAR-Net \cite{sadeghian2018car} & 0.36  & 4.90  					&  23     & 61	 & 158 		& \textbf{2.84}  & \textbf{3.07}  & 4.06 		\\ 
			\midrule
			\ourmodelshort{} (Ours)    			& \textbf{0.25} & \textbf{2.22}     &  \textbf{22}    & \textbf{58} 	 &\textbf{145} 	  & 2.99  & 3.12	&\textbf{3.96}		\\ 
		
			\bottomrule
			\end{tabular}
            \begin{tablenotes}
                \scriptsize
				\item Legend: D=Detector(PIXOR), T=Tracker(UKF+Hungarian), S-LSTM=SocialLSTM, CSP=Convolutional Social Pooling, E2E=End-to-End, Col=Collision rate
			\end{tablenotes}
		\end{threeparttable}
    }
    \vspace{-0.1cm}
	\caption{\textbf{[\nuscenes{}] Social interaction and motion forecasting metrics} at 60\% recall}
	\label{table:nuscenes}
\end{table}

\paragraph{Comparison Against the State-of-the-art}
We benchmark our method against a variety of baselines, which can be classified in three groups.
(i) Methods that use past trajectory as their main motion cues: SocialLSTM \cite{alahi2016social}, Convolutional Social Pooling \cite{deo2018convolutional}, CAR-Net \cite{sadeghian2018car}. 
However, these approaches assume past tracks of every object are given. Thus, we employ our object detector and a vehicle tracker consisting of an Interactive Multiple Model \cite{genovese2001interacting} with Unscented Kalman Filter \cite{wan2000unscented} and Hungarian Matching to extract past trajectories.
(ii) Previously proposed joint detection and motion forecasting approaches: 
FaF \cite{luo2018fast}, IntentNet \cite{casas2018intentnet} and NMP \cite{zeng2019end}. 
(iii) End-to-end (E2E) trainable extensions of the methods in the first group where
  we replace their past trajectory encoders by the actor features coming from our backbone network.

The detection precision-recall (PR) curve for IoU 0.7 
and the accumulated collision rate into the future, for all possible recall points, are shown in Fig. \ref{fig:main_fig} for the \ourdataset{} dataset.
Tables \ref{table:main_table} and \ref{table:nuscenes} show the cumulative collision rate, the centroid L2 error and the absolute heading error for different future horizons
for the \ourdataset{} and \nuscenes{} datasets respectively. 
The results reveal a clear improvement in detection and interaction understanding, as well as a solid gain in long-term motion forecasting. 
Our model substantially improves both in collision rate and centroid error on both datasets. We obtain the lowest heading error on \ourdataset{} while we get results that are on par with the best baseline in \nuscenes{}.
Note that we perform this comparison at 80\% recall at IoU 0.5 in \ourdataset{} because some of the baselines do not reach higher recalls. However, we conduct ablation studies for our model at 95\% recall in the next sections, which is closer to the demand that self-driving cars should meet. In \nuscenes{} we use 60\% recall because all models exhibit worse PR curves, most likely due to the limited dataset size, the sparser LiDAR and the absence of ground-height information.

\begin{table}[t]
      \centering
      \resizebox{\columnwidth}{!}{%
      \small
      \begingroup
      \setlength{\tabcolsep}{5pt} %
      \begin{threeparttable}
      \begin{tabular}{c @{\hspace{1.3\tabcolsep}} c @{\hspace{1.3\tabcolsep}} c @{\hspace{1.3\tabcolsep}} c | c | c @{\hspace{1.5\tabcolsep}} c @{\hspace{1.5\tabcolsep}} c | c @{\hspace{1.5\tabcolsep}} c @{\hspace{1.5\tabcolsep}} c }
      \toprule
      Decoder   & R & B & T      & C (\permil)        & \multicolumn{3}{c}{Centroid @ 3s}      & \multicolumn{3}{c}{Heading @ 3s}                     \\
       & & &                              & 0-3s          & L2            & NLL           & H         & L2       & NLL      & H            \\
      \midrule
      MLP & \xmark  & \xmark  & \xmark    & 9.79          & 127           & 2.56          & 0.47      & 5.68          &  -3.32          &  0.15               \\
      MLP & \cmark  & \xmark  & \xmark    & 2.23          & 118           & 1.50          & 0.47      & 4.97          &  -7.01          &  1.49               \\
      \midrule
      GNN & \cmark  & \xmark  & \xmark    & 2.91          & 122           & 1.73          & 0.72      & 5.09          &  -6.65          &  2.37               \\
      GNN & \cmark  & G       & \xmark    & 2.14          & 116           & 1.42          & 0.50      & 5.14          &  \textbf{-7.31} &  1.17               \\
      GNN & \cmark  & R       & \xmark    & 1.32          & 109           & 1.14          & 0.39      & 4.77          &  -7.12          &  -1.97              \\
      GNN & \cmark  & R       & R         & \textbf{0.78} & \textbf{105}  & \textbf{1.08} & 0.24      & \textbf{4.75} &  -6.99          &  -1.87              \\
      \bottomrule
      \end{tabular}
      \begin{tablenotes}
        \scriptsize
        \item Legend: R=RRoI, B=bounding box, Traj=future trajectory, C=Collision rate
      \end{tablenotes}
      \end{threeparttable}
      \endgroup
      }
      \vspace{-0.1cm}
      \caption{\textbf{[\ourdataset] Ablation study of the different contributions} at 95\% recall}
      \vspace{-0.3cm}  
      \label{table:ablation}
  \end{table}

\paragraph{Ablation Study} 
We first study the per-actor feature extraction mechanism of our model by comparing simple feature indexing versus Rotated RoI Align. 
We do this study in a model that does not contemplate interactions for better isolation. In particular, we omit our \ourmodelshort{} from Fig. \ref{fig:architecture}
and use the initial trajectories predicted by $\operatorname{CNN_R}$ as the final output.
As shown in the first 2 rows of Table \ref{table:ablation}, 
 Rotated RoI Align clearly outperforms simple feature indexing. 
Our RRoI pooled features provide better information about the surroundings of the target vehicle since they contain  features in a region spanning by 25 meters
whereas the feature indexing variant consists of just accessing the feature map at the anchor location associated to the detection.

\paragraph{Graph Neural Network architectures} We evaluate several GNN architectures with different levels of spatial awareness to demonstrate the effectiveness of our \ourmodelshort{}.
The second to third rows of Table \ref{table:ablation} show that adding a standard GNN without spatial awareness does not improve performance. 
From the third to the forth row we observe an improvement by including the detection bounding boxes in global (G) coordinate frame 
as part of the state at every message passing iteration of the GNN. 
Then, we make these bounding boxes relative (R) to the actor receiving the message, which gives us a boost across most metrics.
Finally, we add the parameters of the predicted probability distributions of the future trajectories to the message passing algorithm to recover our \ourmodelshort{}. 
Interestingly, the models become more certain, i.e. lower entropy (H), as we add better spatial-awareness mechanisms.

\begin{figure}
        \centering
        \resizebox{\columnwidth}{!}{%
        \includegraphics[page=1, width=\textwidth, trim={0.25cm 3.5cm 0.25cm 1.9cm}, clip]{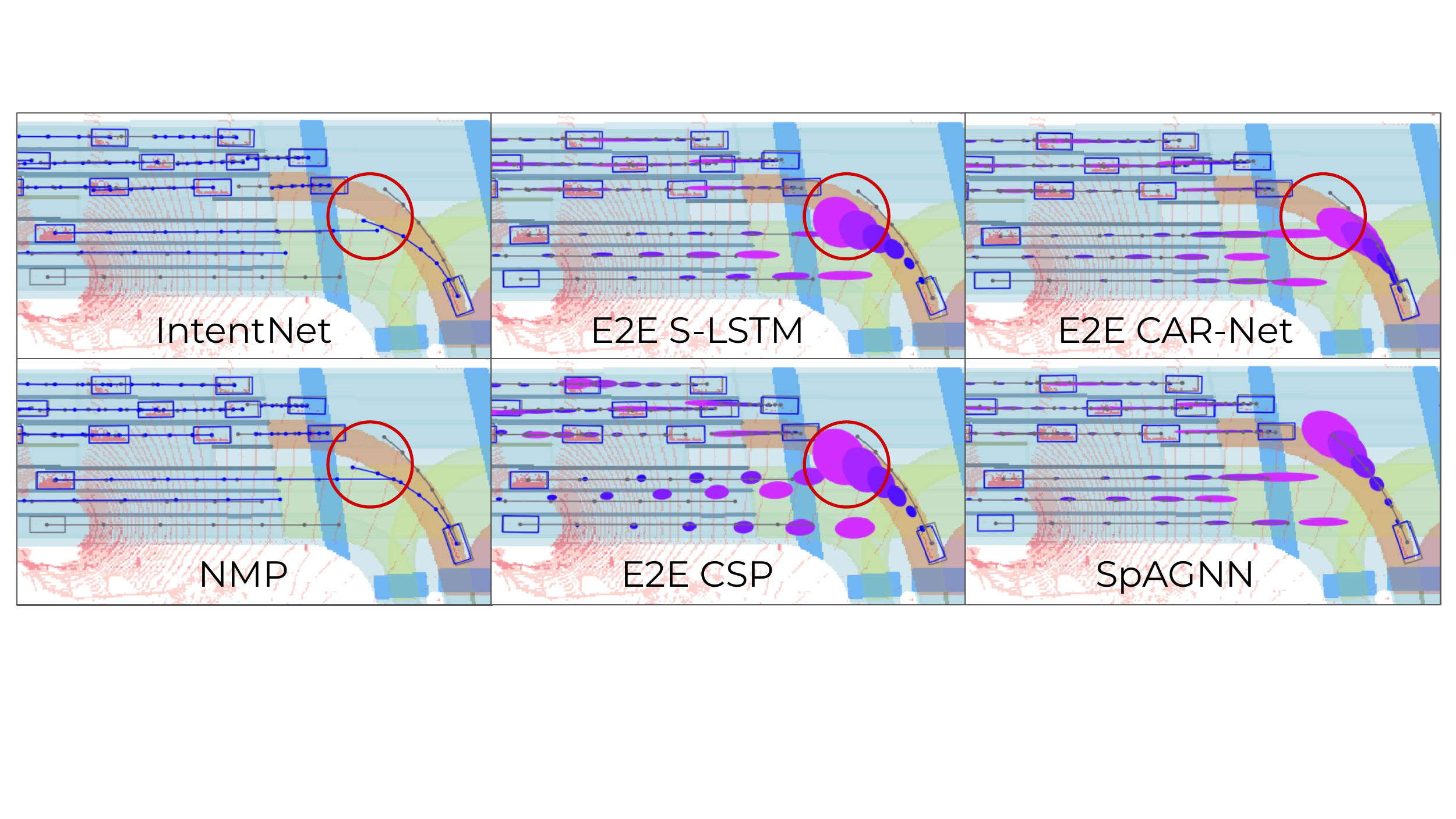}
        }
      \caption{\textbf{Qualitative results.} Red circles denote collisions}
        \label{fig:qualitative}
        \vspace{-0.5cm}  
  \end{figure}

\paragraph{Qualitative results} Fig. \ref{fig:qualitative} shows the outputs from the baselines and our \ourmodelshort{} in a crowded scene in the \ourdataset{} dataset. More visualizations will be added to our video submission. 
The waypoint trajectories output by IntentNet and NMP are shown in blue (they do not model uncertainty).
The bivariate Gaussian distribution output by the rest of the models is shown in a colormap to encode time: from blue at 0 seconds to pink at 3 seconds into the future. 
The principal components of the ellipsis correspond to the square root of the two eigenvectors of the predicted covariance matrix.
The ground-truth trajectories are drawn in gray. 
This example illustrates a failure across all the baselines, which predict that a pair of vehicles are going to collide, thus failing to model the agent-agent interactions. 
Note that the relational baselines and \ourmodelshort{} also predict the future states of the vehicle that performed the data collection (see the epicenter of the LiDAR point cloud) since it plays an important role in the social interactions.

\section{Conclusion}
\vspace{-0.1cm}

In this paper we  tackled the problem of detection and relational behavior forecasting. 
Unlike existing approaches, we have proposed a single model that can reason jointly about these tasks. 
We have designed a novel spatially-aware graph neural network to produce socially coherent, probabilistic estimates of future trajectories. 
Our approach resulted in significant improvements over the state-of-the-art on the challenging \ourdataset{} and \nuscenes{} autonomous driving datasets. 
We plan to extend our model to generate multiple future outcomes of the scene, to use other sensory input such as images and radar, and to reason about other types of agents such as pedestrians and cyclists.

\pagebreak

\bibliography{root}

\section{Appendix}

\subsection{Algorithm}

The full inference algorithm is described in Alg. \ref{alg:spagnn}. We write the algorithm in a non-vectorized manner for the sake of readability, although our implementation is fully vectorized since inference time is critical in autonomy tasks onboard a self-driving car. The proposed algorithm would run every 0.1 seconds (when a new LiDAR sweep is gathered). We avoided adding an outer loop to Alg. \ref{alg:spagnn} reflecting this in order to avoid cluttering the notation.

\begin{algorithm*}
\caption{\ourmodelshort{}} \label{alg:spagnn}
\textbf{Input:} 
Multiple LiDAR sweeps compensated by ego-motion $L$ and raster map $M$: $\Omega=\begin{Bmatrix} L, M \end{Bmatrix}$.
Number of message passing steps of GNN $K$.

\textbf{Output:} Detections (confidences and bounding boxes) $D = \begin{Bmatrix} c_0,b_0, c_1,b_1, ..., c_N,b_N \end{Bmatrix}$. Prediction of statistics $O^K = \begin{Bmatrix} o_0^K, o_1^K, ..., o_N^K \end{Bmatrix}$

\begin{algorithmic}[1]
\State $ \begin{Bmatrix} c_0,b_0, c_1,b_1, ..., c_N,b_N \end{Bmatrix} \gets \operatorname{Detector}\left(\Omega \right)$ \Comment{Includes NMS and confidence thresholding}
\For{$i=1, ..., N$} \Comment{Compute actor initial features and trajectories independently}
    \State $r_i \gets \operatorname{RRoiAlign}(b_i)$ 
    \State $h_i^0 \gets \operatorname{MaxPooling}(\operatorname{CNN}(r_i))$ 
    \State $o_i^0 \gets \operatorname{MLP}(h_i^0)$ 
\EndFor
\State Construct $G = (V, E)$ a graph from detections
\State Compute pairwise cooridinate transformations $\mathcal{T}_{u,v}$, $\forall (u, v) \in E $
\For{$k=1, ..., K$} \Comment{Loop over graph propagations}
	\For {$ (u, v) \in E $} \Comment{Compute message for every edge in the graph}
		\State $m_{u \rightarrow v}^{(k)} = \mathcal{E}^{(k)}\left(h_{u}^{k-1}, h_{v}^{k-1}, \mathcal{T}_{u,v} (o_{u}^{(k-1)}), o_{v}^{(k-1)}, b_{u}, b_{v} \right)$
	\EndFor
	\For {$ v \in V $} \Comment{Update node states}
		\State $a_{v}^{(k)}=\mathcal{A}^{(k)}\left(\left\{m_{u \rightarrow v}^{(k)} : u \in \mathbf{N}(v)\right\}\right)$ \Comment{Aggregate messages from neighbors}
		\State $h_{v}^{(k)}=\mathcal{U}^{(k)}\left(h_{v}^{(k-1)}, a_{v}^{(k)}\right)$ \Comment{Update the hidden state}
		\State $o_{v}^{(k)}=\mathcal{O}^{(k)}\left(h_{v}^{(k)}\right)$ \Comment{Update the output state (probabilistic trajectory)}
	\EndFor
\EndFor
\State\Return $D, O^K$
\end{algorithmic}
\end{algorithm*}

The reader might wonder why the future time steps do not appear explicitly in Alg. \ref{alg:spagnn}. We note that $h_v^k$ is the hidden state for agent $v$ at the $k$-th iteration of message passing, acting as a summarization of appearance, motion and interaction features from actor $v$ and its neighbors. The states for all future time steps are predicted in a feed-forward fashion, using an MLP, as indicated in line 14 of Alg. \ref{alg:spagnn}. We did not observe any gain by using a recurrent trajectory decoder that explicitly reasons about time dependencies in the future trajectory and therefore we adopted this formulation for simplicity.

\subsection{Implementation details} \label{implementation_details}

In this section we describe our implementation, including details about the network architecture of the different components as well as training.

\paragraph{Detection network} Our LiDAR backbone uses 2, 2, 3, and 6 layers in its 4 residual blocks. 
The convolutions in the residual blocks of our LiDAR backbone have 32, 64, 128 and 256 filters with a stride of 1, 2, 2, 2 respectively. 
The backbone that processes the high-definition maps uses 2, 2, 3, and 3 layers in its 4 residual blocks. 
The convolutions in the residual blocks of our map backbone have 16, 32, 64 and 128 filters with a stride of 1, 2, 2, 2 respectively. 
For both backbones, the final feature map is a multi-resolution concatenation of the outputs of each residual block, as explained in \cite{yang2018pixor}.
This gives us 4x downsampled features with respect to the input.
The header network consists of 4 convolution layers with 256 filters per layer. 
We use GroupNorm \cite{wu2018group} because of our small batch size (number of scenarios) per GPU.

Because we want our detector to have a very high recall, we move away from only targeting to detect cars with at least 1 LiDAR point on its surface in the current frame as was done in previous works \cite{luo2018fast,casas2018intentnet,zeng2019end}.
In contrast, we require that the car either has 1 LiDAR point in the current LiDAR sweep or 2 points in 2 different previous LiDAR sweeps so that the model has some information to figure out its current location based on motion.

Finally, we reduce the candidate confidence scores and bounding box proposals by first taking the top 200 anchors ranked by confidence score and then applying NMS with an IoU threshold of 0.1 in order to obtain our final detections.

\paragraph{Per-actor feature extraction} We use a CUDA kernel for fast Rotated Region of Interest Align (RRoI Align \cite{ma2018arbitrary}).
We use a 41 by 25 meters region of interest aligned to the detection bounding box heading around each target vehicle, with 31 meters in front of the car, 10 meters behind and 12.5 meters to each side. 
We use an output resolution of 1 m/pixel in our RRoI Align operator. 
Our (256, 41, 25)-dimensional tensor per-actor is then downsampled 8 times by a 3-layer CNN, and the features increased to 512. 
Then, we apply max-pooling across the remaining spatial dimensions to obtain the initial hidden state $h_i^0$.
Finally, this is processed by a 2-layer MLP to produce the initial output state $o_i^0$.

\paragraph{Relational Behavior Forecasting} We implement a fully-vectorized version of our \ourmodelshort{}. 
We use $K=3$ propagation steps. The parameters of the edge $\mathcal{E}^{(k)}$, aggregate $\mathcal{A}^{(k)}$, update $\mathcal{U}^{(k)}$ and output $\mathcal{O}^{(k)}$ functions are shared across all propagation steps since we did not observe any improvements by having separate ones. 
This highlights the refinement nature of our probabilistic trajectory prediction process.
Our edge function $\mathcal{E}^{(k)}$ consists of a 3-layer MLP that takes as input the hidden states of the 2 terminal nodes at each edge in the graph at the previous propagation step 
as well as the projected output states at the previous iteration, processed by a 2-layer MLP.
We use feature-wise max-pooling as our aggregate function $\mathcal{A}^{(k)}$ in order to be more robust to changes in the graph topology. 
We use an efficient CUDA kernel to implement the scatter-max operation that receives the incoming messages from neighboring nodes and outputs the aggregated message.
To update our hidden states we use a GRU cell as $\mathcal{U}^{(k)}$. Finally, to output the statistics of our multivariate Gaussian and Von Mises distribution we use a 2-layer MLP $\mathcal{O}^{(k)}$.

\paragraph{Scheduled sampling} Our end-to-end learnable model first detects the vehicles in the scene and then forecasts its motion for the future 3 seconds.
We recall that the state updates in our \ourmodelshort{} are dependent on the incoming messages from neighboring nodes/vehicles. Thus, we add scheduled sampling \cite{he2017mask} during training in order to mitigate the distribution mismatch between ground-truth bounding boxes and detected bounding boxes. 
More precisely, we start training the second stage of our model by feeding only the ground-truth boxes since at that stage the detection is not good enough and the big amount of false positives and false negatives will mislead the learning of the GNN parameters. As detection gets better, we increase the probability of replacing the ground-truth bounding boxes with detections. 
In practice, we start with probability 1.0 of using ground-truth bounding boxes, lower it to 0.7 after 10,000 training iterations and finally down to 0.3 after 20,000. This leads to improved results because we avoid confusing the GNN with false positive and false negative detections while the detector is still at an early learning phase.

\paragraph{Optimizer} We use Adam \cite{kingma2014adam} optimizer with a base learning rate of $1.25e-5$, which gets linearly increased with the batch size.
In our experiments we use a batch size of 3 per GPU and a total of 4 GPUs for a total batch size of 12, giving us a final learning rate of $1.5e-4$. 

\subsection{Baselines details} \label{baselines_details}

\paragraph{Tracking-based baselines} To keep the same architectures that were proposed in SocialLSTM \cite{alahi2016social}, Convolutional Social Pooling \cite{deo2018convolutional}, and CAR-Net \cite{sadeghian2018car}, past trajectories are needed. However, our proposed model is tracking-free and just takes the sensor and map data as input. Thus, we implement an Interactive Multiple Model \cite{genovese2001interacting} with Unscented Kalman Filter \cite{wan2000unscented} and Hungarian Matching to extract past trajectories. These past trajectories are used to obtain our results for D+T+S-LSTM, D+T+CSP and D+T+CAR-Net in Table \ref{table:main_table}. In particular, we feed up to 1 second of past trajectory. The following tracker settings delivered the best performance: filter out detections with a confidence score lower than 0.5 (this still delivers a detection recall higher than 95\% in \ourdataset{}), wait 9 cycles before discarding a track completely, birth a track immediately if the confidence score of a detection is higher than 0.9 and require a minimum of 3 consecutive IoU-based matches to birth a track otherwise. We tried training with past ground-truth trajectories as well as with tracking results. Directly training with tracking results delivered slightly better results, which corresponds to the numbers reported in Table \ref{table:main_table}. 

\paragraph{End-to-end adapted baselines} The tracking-based baselines do not achieve a good performance partly due to its hard dependency on the tracking quality. 
Therefore, we seek to benchmark previously proposed interaction operators and trajectory decoders and \ourmodelshort{} in a better isolated experiment. In order to do this, we keep our backbone network and feature extraction and replace the second stage in our implementation by the interaction operators and trajectory decoders proposed in SocialLSTM \cite{alahi2016social}, Convolutional Social Pooling \cite{deo2018convolutional}, and CAR-Net \cite{sadeghian2018car}. However, because the proposed methods use past trajectories and we do not, we need to adapt their trajectory encoders. 
For E2E S-LSTM, instead of unrolling a SocialLSTM all the way from past trajectory to the future trajectory, we feed the per-actor features extracted by our backbone network fed as input to a SocialLSTM that unrolls from the current time to the future.
For E2E CSP, the LSTM encoder is removed from Convolutional Social Pooling and the social tensor is directly initialized with the per-actor features extracted by our backbone network. 
For E2E CAR-Net, we use our backbone with RRoI pooling to replace its feature extractor module and the per-actor features as the past motion context. 

\paragraph{Joint Perception and Prediction baselines} FaF \cite{luo2018fast}, IntentNet \cite{casas2018intentnet} and NMP \cite{zeng2019end} did not precise any adaptation.

\subsection{Additional qualitative results}
\begin{figure*}[b]
\centering
\begin{subfigure}{.5\textwidth}
    \centering
    \includegraphics[width=0.95\textwidth, trim={3.5cm 3.0cm 3.5cm 3.0cm}, clip]{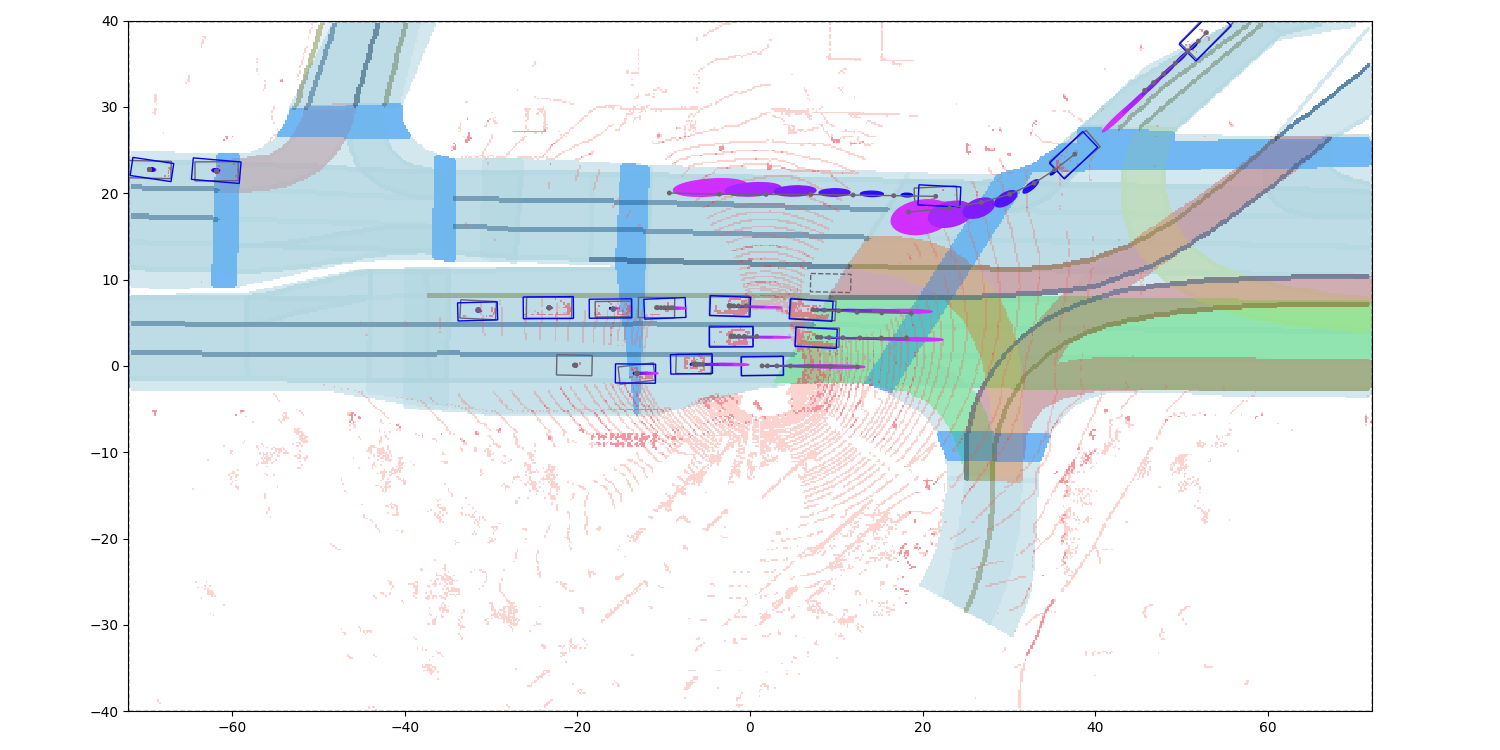}
\end{subfigure}%
\begin{subfigure}{.5\textwidth}
    \centering
    \includegraphics[width=0.95\textwidth, trim={3.5cm 3.0cm 3.5cm 3.0cm}, clip]{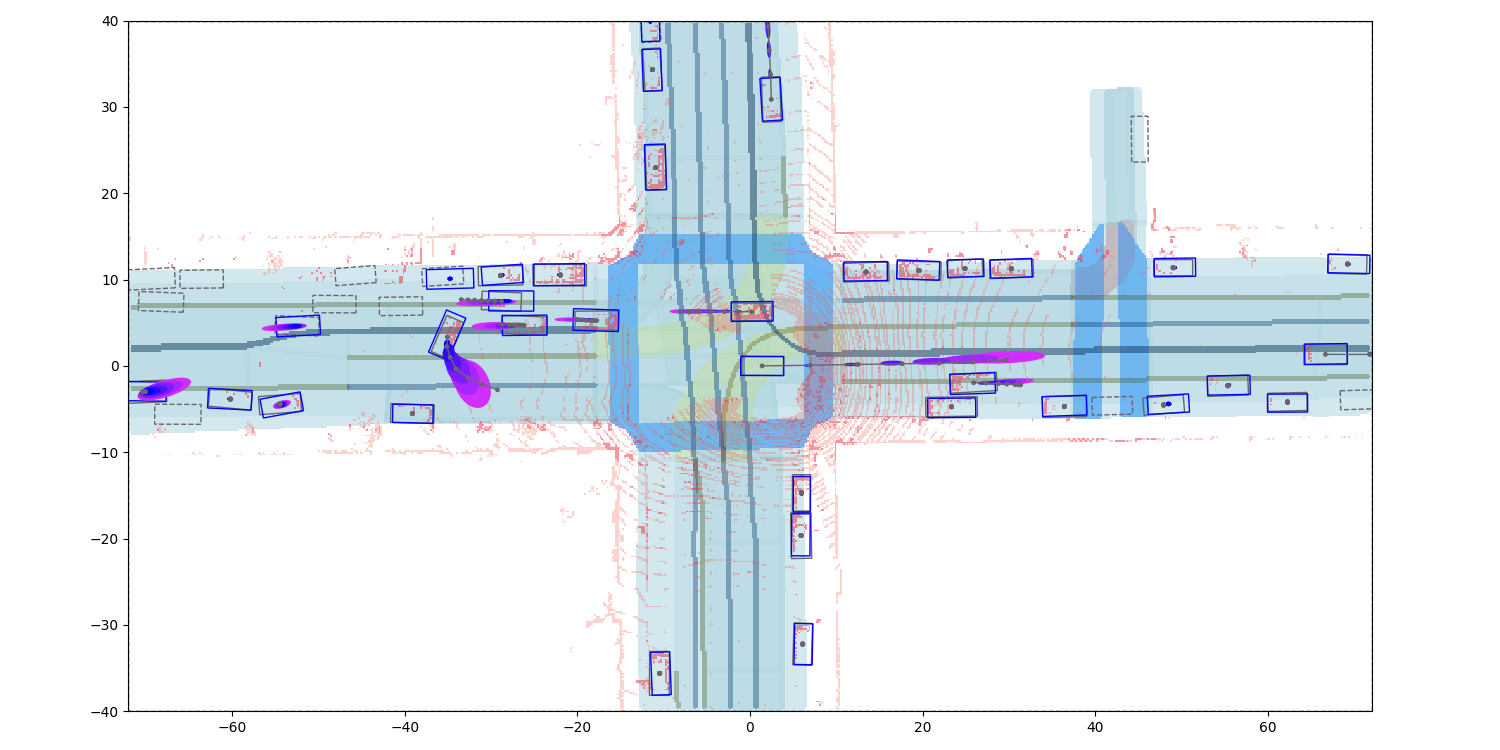}
\end{subfigure}
\vspace{0.2cm}
\begin{subfigure}{.5\textwidth}
    \centering
    \includegraphics[width=0.95\textwidth, trim={3.5cm 3.0cm 3.5cm 3.0cm}, clip]{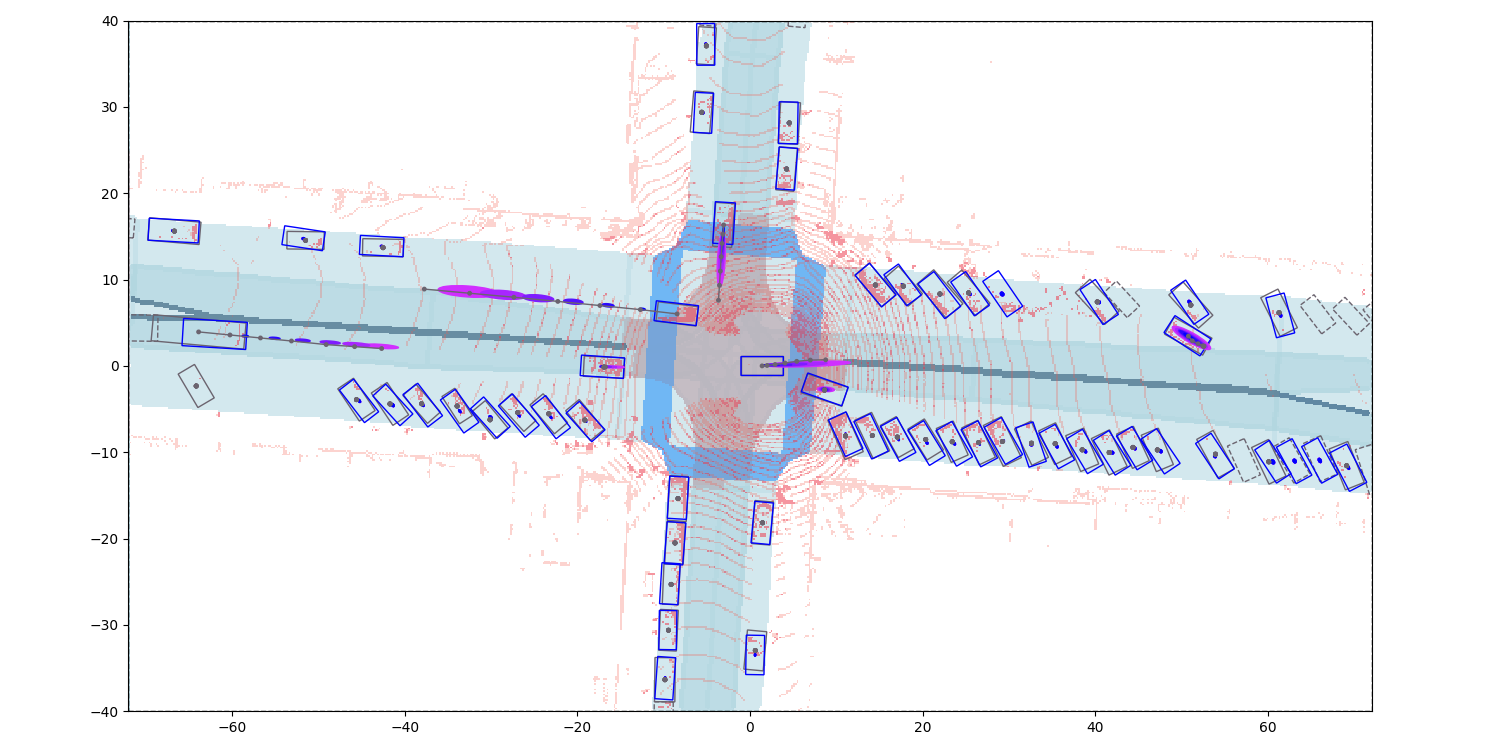}
\end{subfigure}%
\begin{subfigure}{.5\textwidth}
    \centering
    \includegraphics[width=0.95\textwidth, trim={3.5cm 3.0cm 3.5cm 3.0cm}, clip]{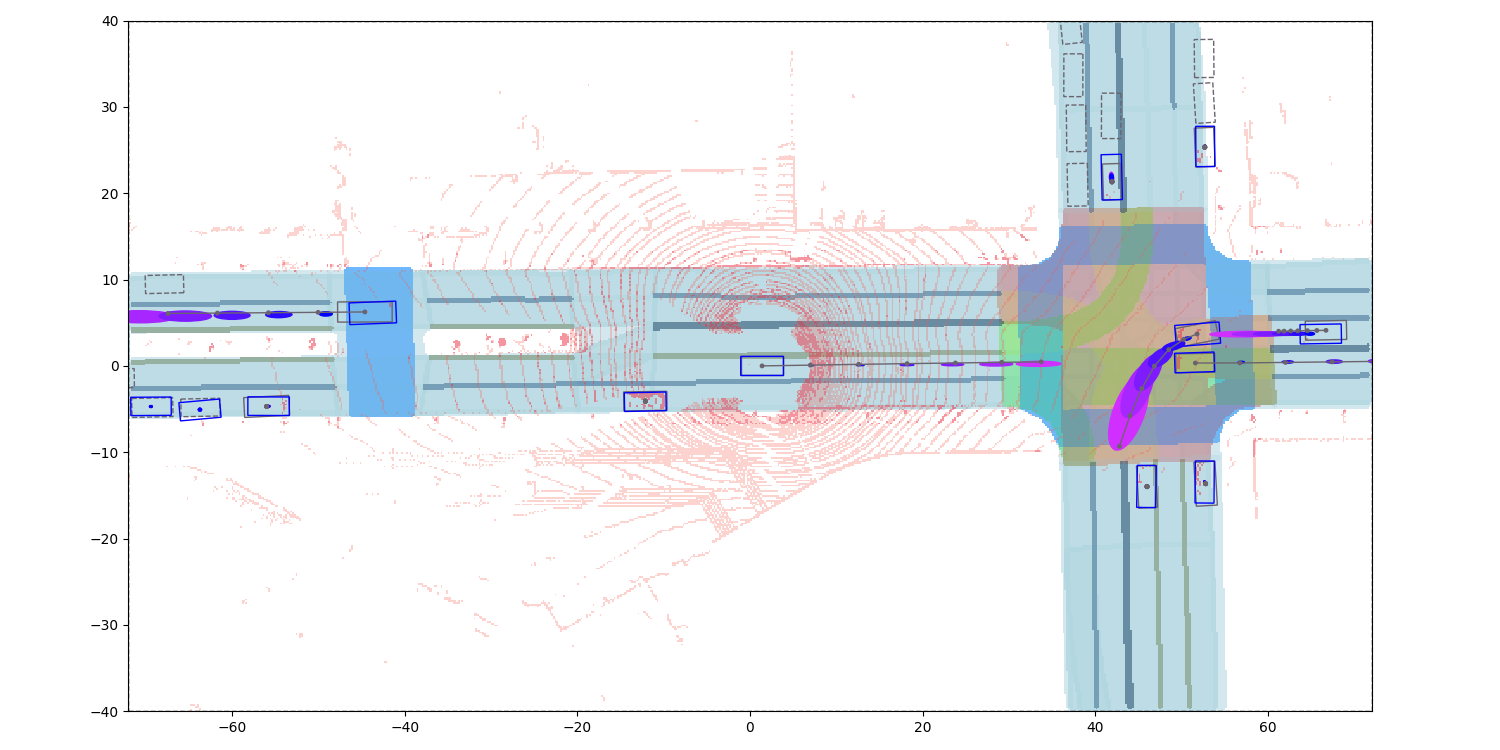}
\end{subfigure}
\vspace{0.2cm}
\begin{subfigure}{.5\textwidth}
    \centering
    \includegraphics[width=0.95\textwidth, trim={3.5cm 3.0cm 3.5cm 3.0cm}, clip]{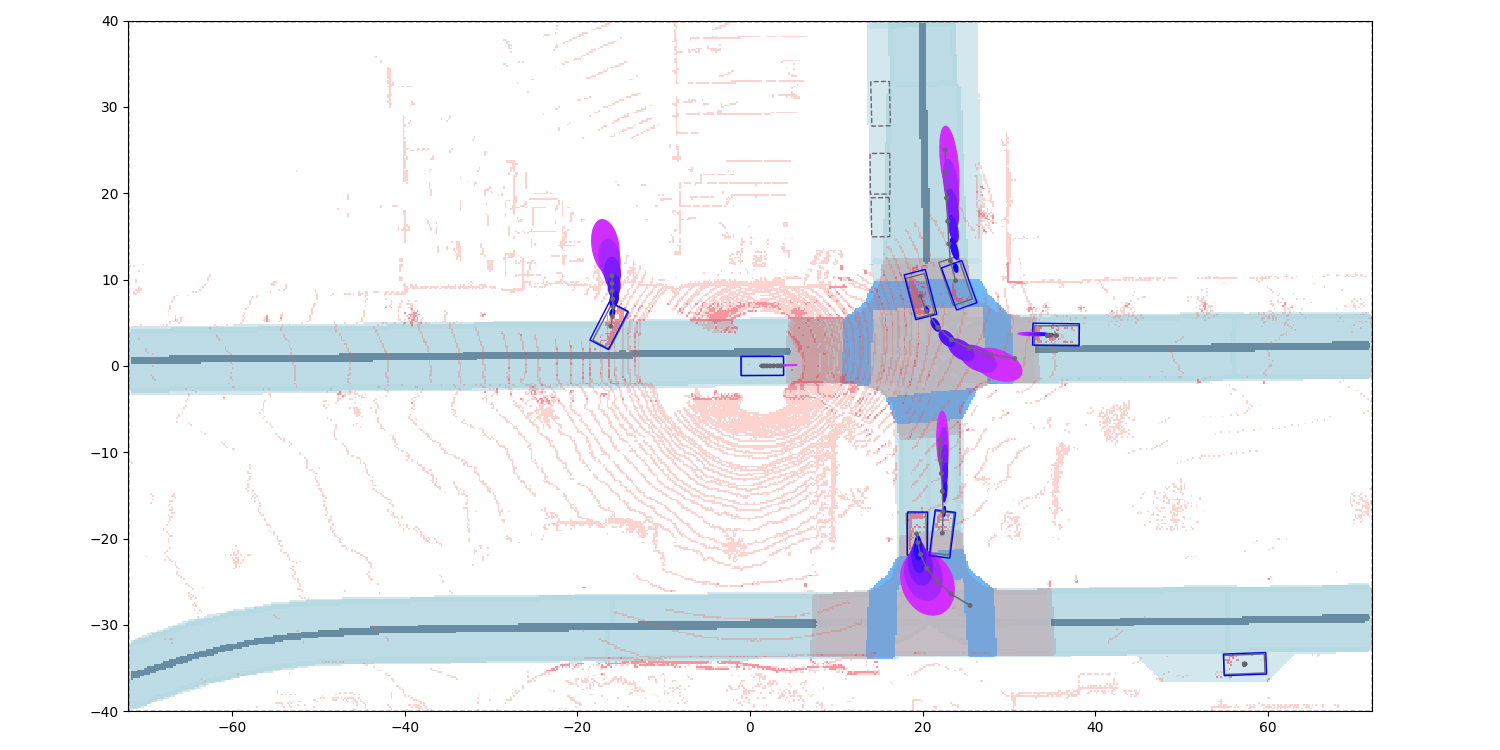}
\end{subfigure}%
\begin{subfigure}{.5\textwidth}
    \centering
    \includegraphics[width=0.95\textwidth, trim={3.5cm 3.0cm 3.5cm 3.0cm}, clip]{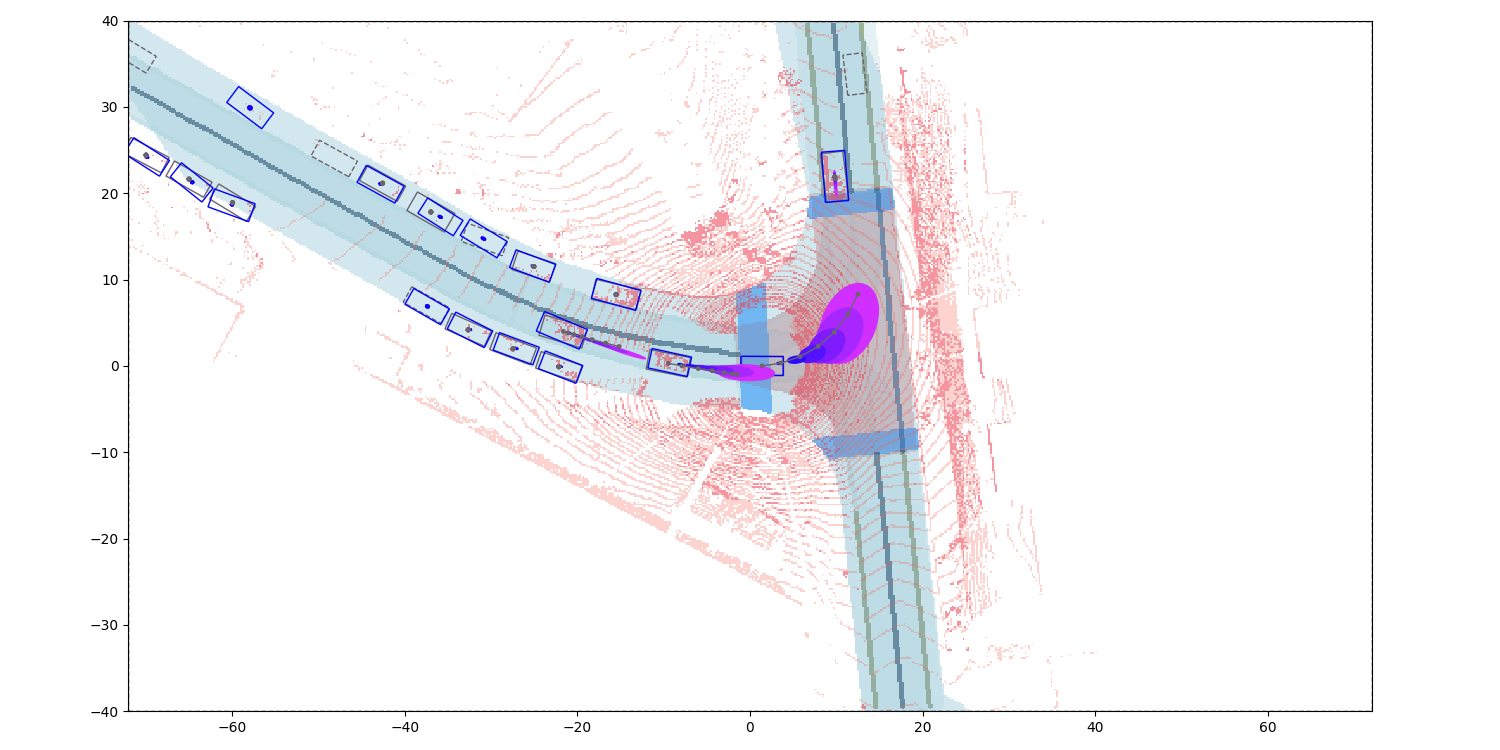}
\end{subfigure}
\vspace{0.2cm}
\begin{subfigure}{.5\textwidth}
    \centering
    \includegraphics[width=0.95\textwidth, trim={3.5cm 3.0cm 3.5cm 3.0cm}, clip]{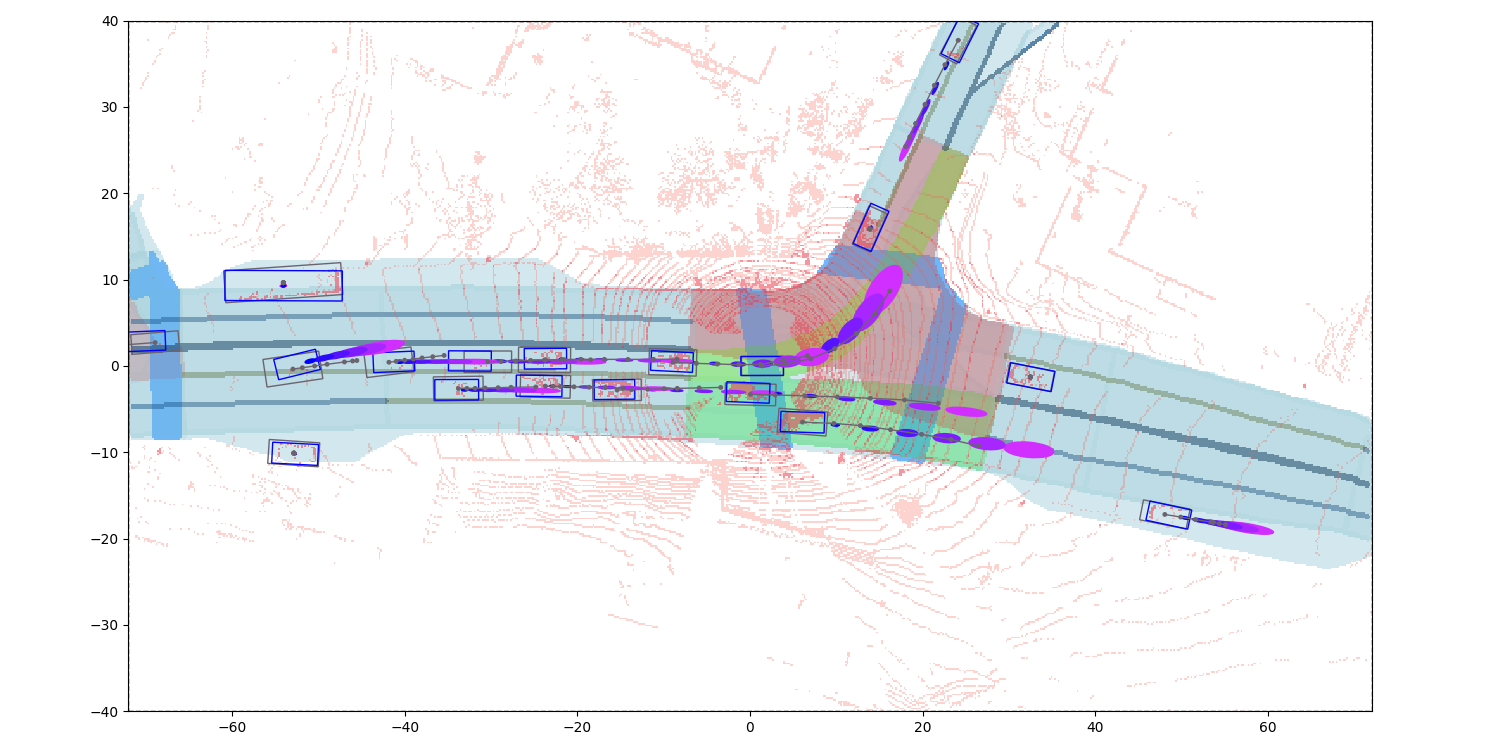}
\end{subfigure}%
\begin{subfigure}{.5\textwidth}
    \centering
    \includegraphics[width=0.95\textwidth, trim={3.5cm 3.0cm 3.5cm 3.0cm}, clip]{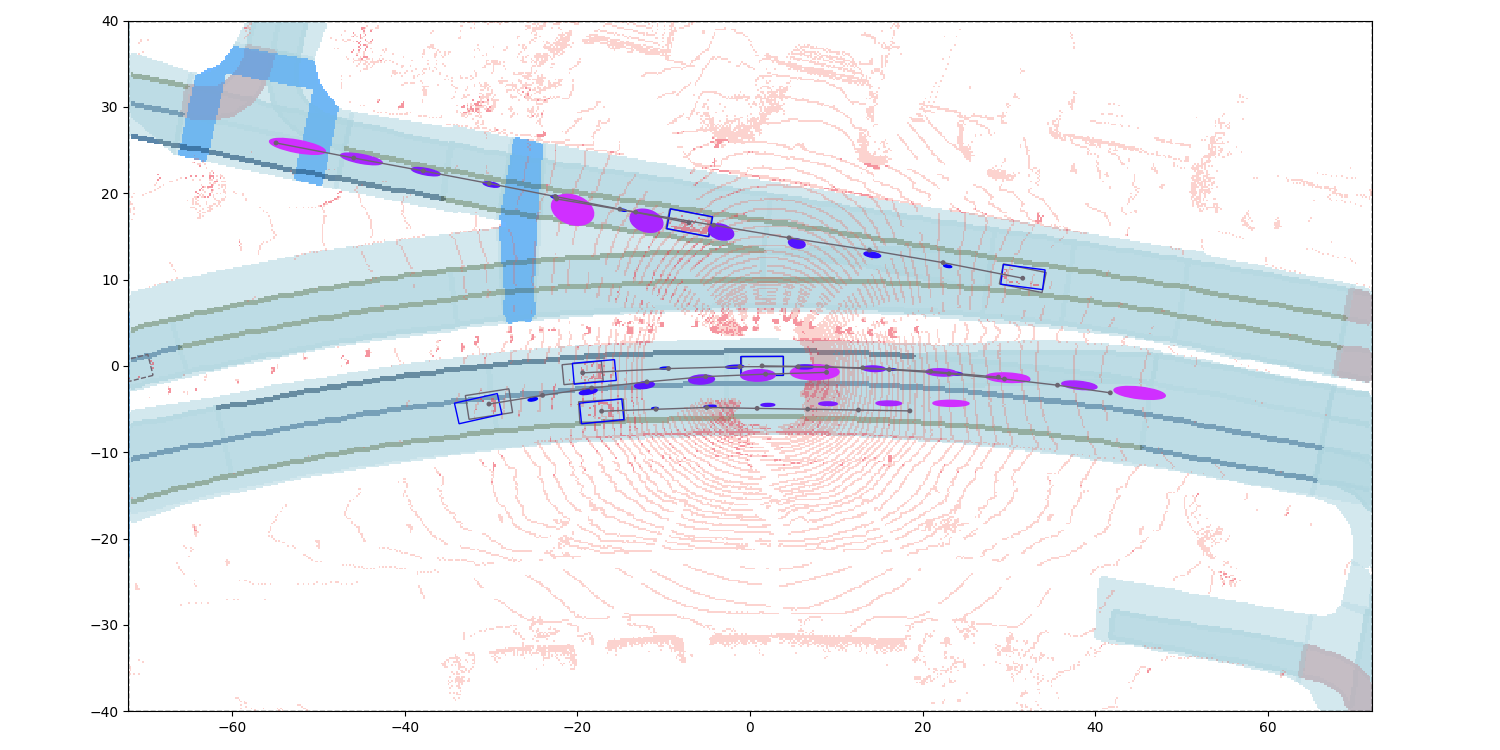}
\end{subfigure}
\vspace{0.2cm}
\begin{subfigure}{.5\textwidth}
    \centering
    \includegraphics[width=0.95\textwidth, trim={3.5cm 3.0cm 3.5cm 3.0cm}, clip]{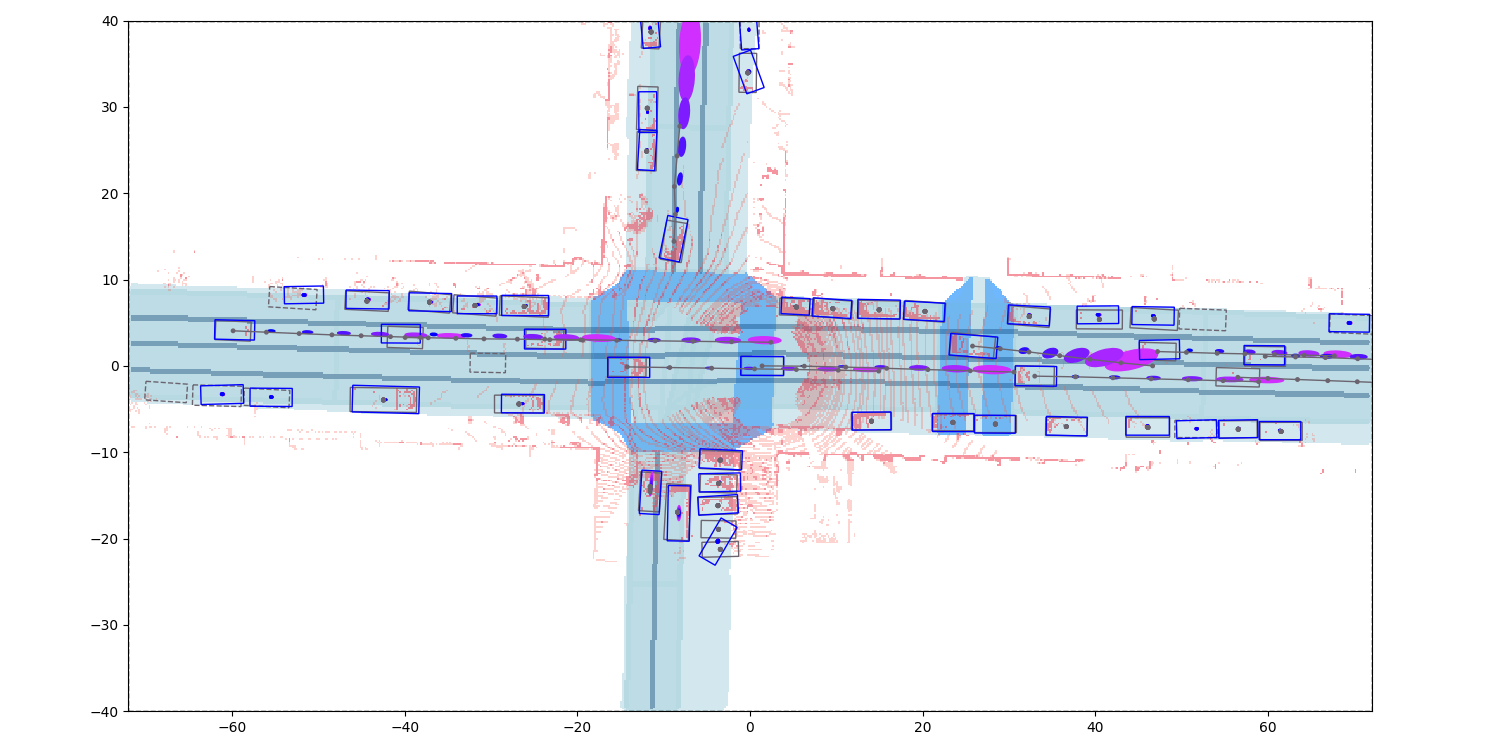}
\end{subfigure}%
\begin{subfigure}{.5\textwidth}
    \centering
    \includegraphics[width=0.95\textwidth, trim={3.5cm 3.0cm 3.5cm 3.0cm}, clip]{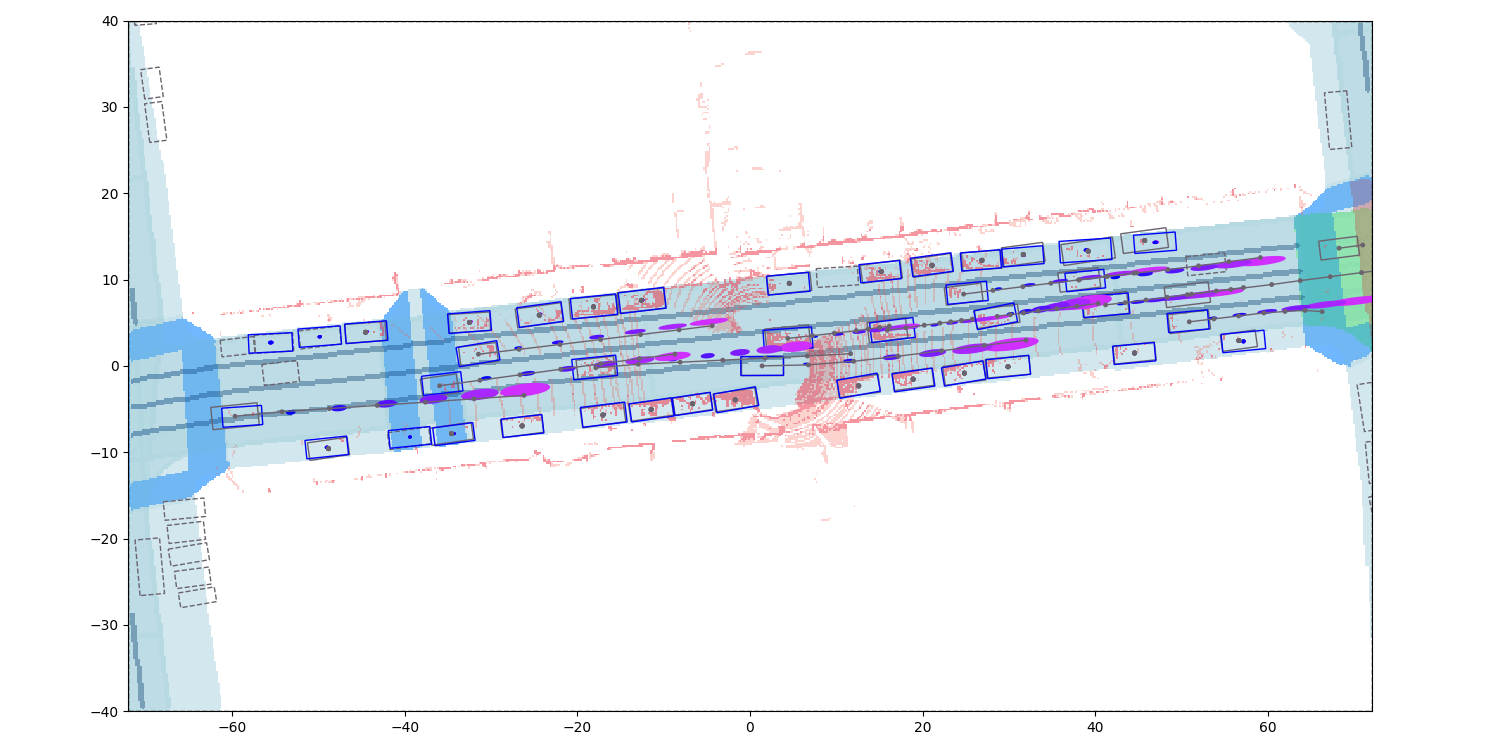}
\end{subfigure}
\vspace{0.2cm}
\begin{subfigure}{.5\textwidth}
    \centering
    \includegraphics[width=0.95\textwidth, trim={3.5cm 3.0cm 3.5cm 3.0cm}, clip]{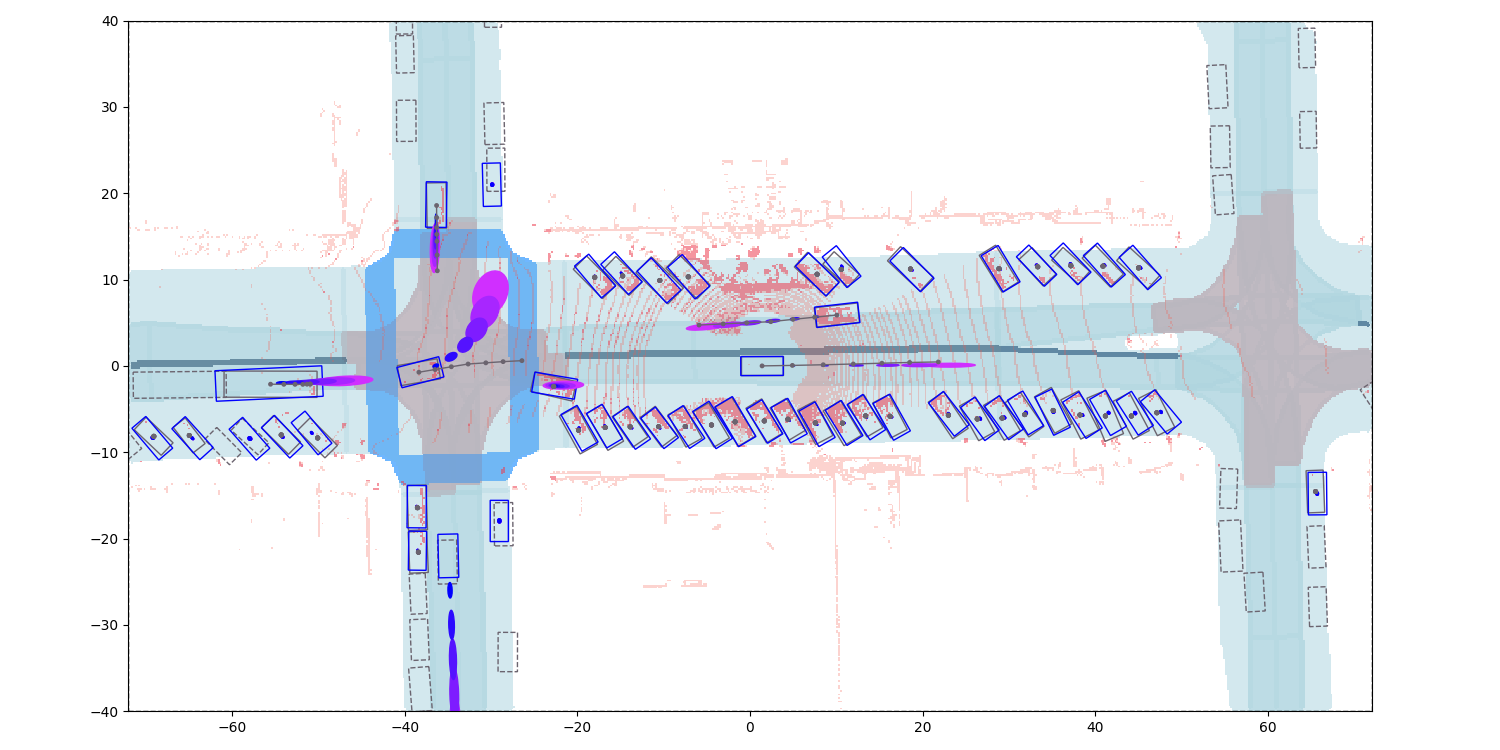}
\end{subfigure}%
\begin{subfigure}{.5\textwidth}
    \centering
    \includegraphics[width=0.95\textwidth, trim={3.5cm 3.0cm 3.5cm 3.0cm}, clip]{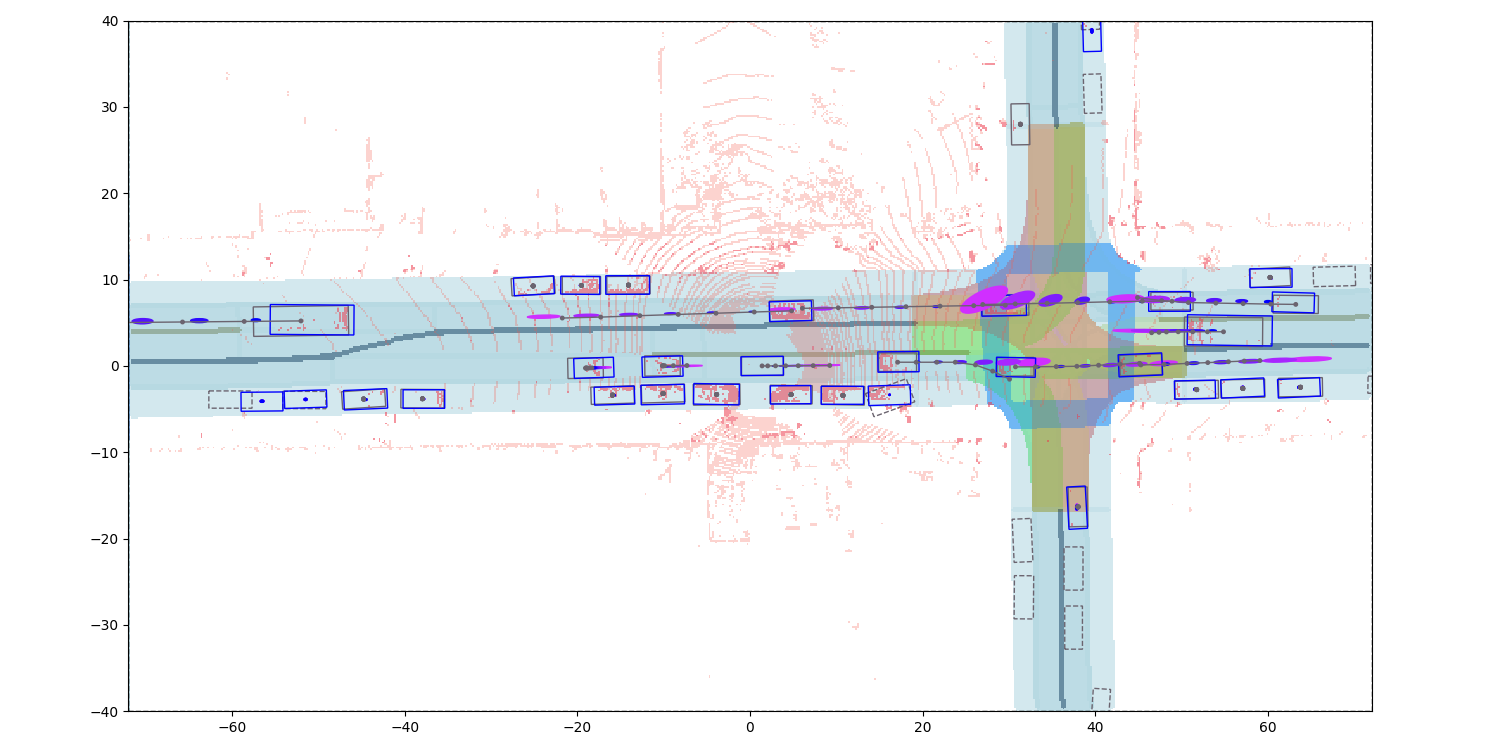}
\end{subfigure}
\caption[short]{\textbf{[\ourdataset{}] Additional qualitative results} of our \ourmodelshort{}. Detections are shown as blue bounding boxes. Probabilistic motion forecasts are shown as ellipsis (corresponding to one standard deviation of a bivariate gaussian) where variations in color indicate different future time horizons (from 0 seconds in blue to 3 seconds in pink).
Ground-truth boxes and future waypoints are displayed in gray. A dashed gray box means the object is occluded. Last row shows failure modes.}
\label{fig:additional_qualitative}
\end{figure*}

\begin{figure*}[b]
\centering
\begin{subfigure}{.33\textwidth}
    \centering
    \includegraphics[width=0.97\textwidth, trim={12cm 3.0cm 12cm 3.0cm}, clip]{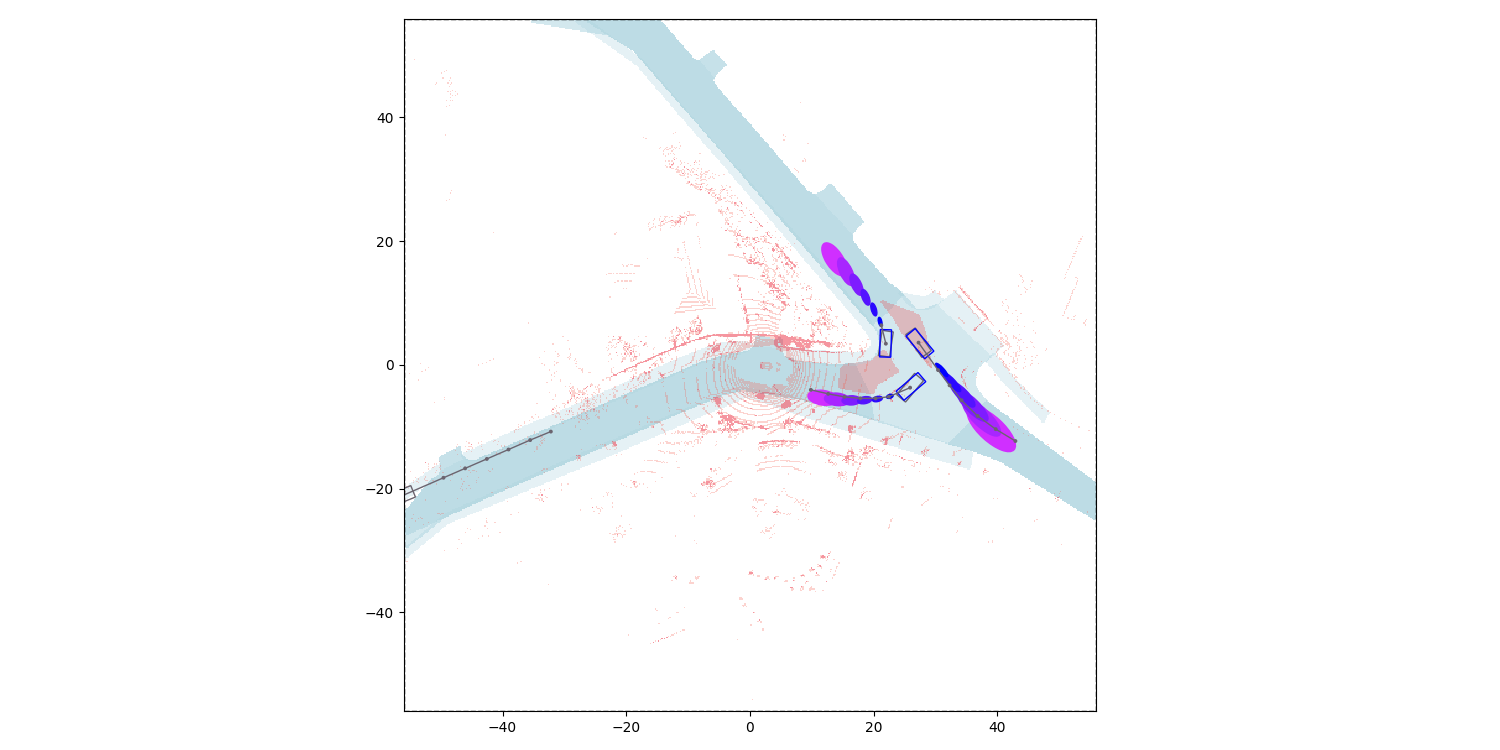}
\end{subfigure}%
\begin{subfigure}{.33\textwidth}
    \centering
    \includegraphics[width=0.97\textwidth, trim={12cm 3.0cm 12cm 3.0cm}, clip]{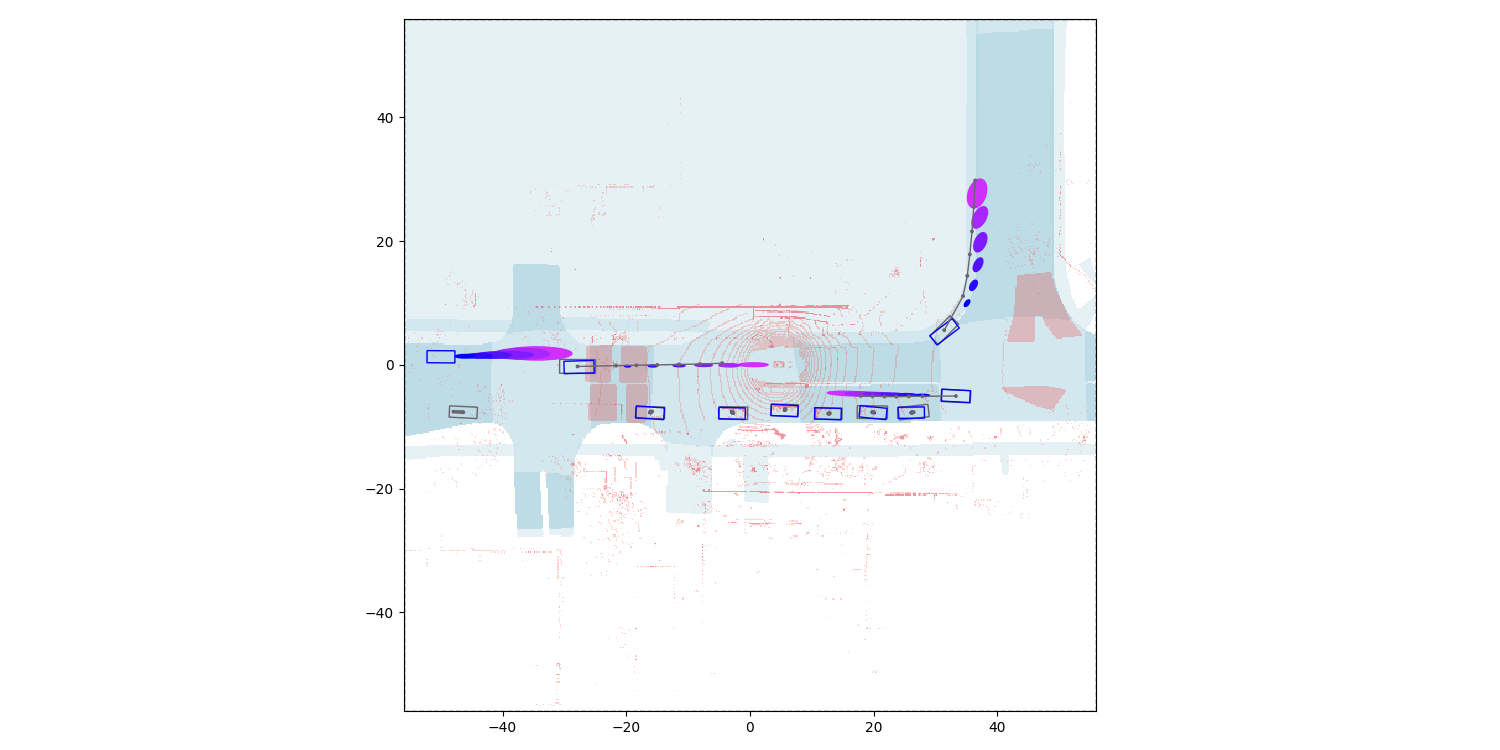}
\end{subfigure}%
\begin{subfigure}{.33\textwidth}
    \centering
    \includegraphics[width=0.97\textwidth, trim={12cm 3.0cm 12cm 3.0cm}, clip]{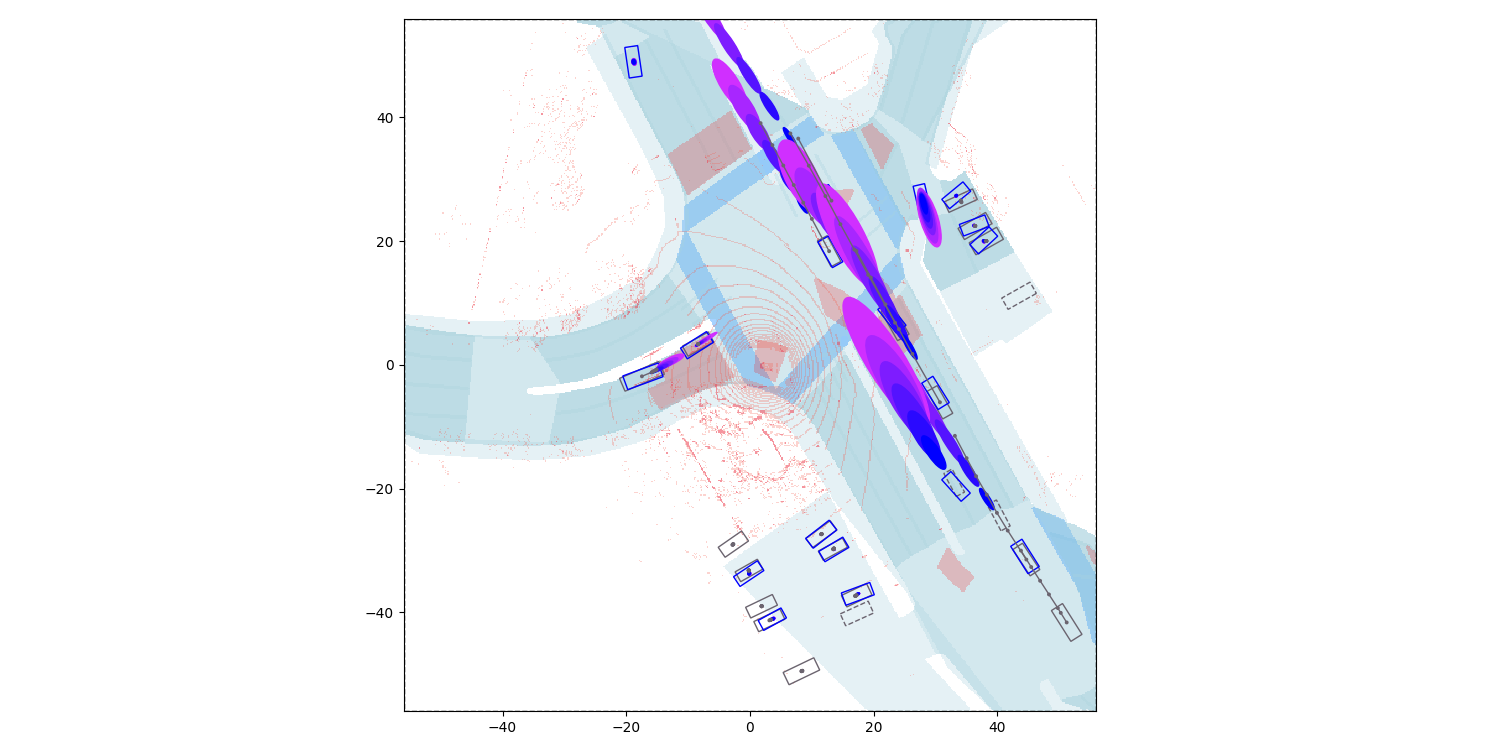}
\end{subfigure}
\vspace{0.2cm}
\begin{subfigure}{.33\textwidth}
    \centering
    \includegraphics[width=0.97\textwidth, trim={12cm 3.0cm 12cm 3.0cm}, clip]{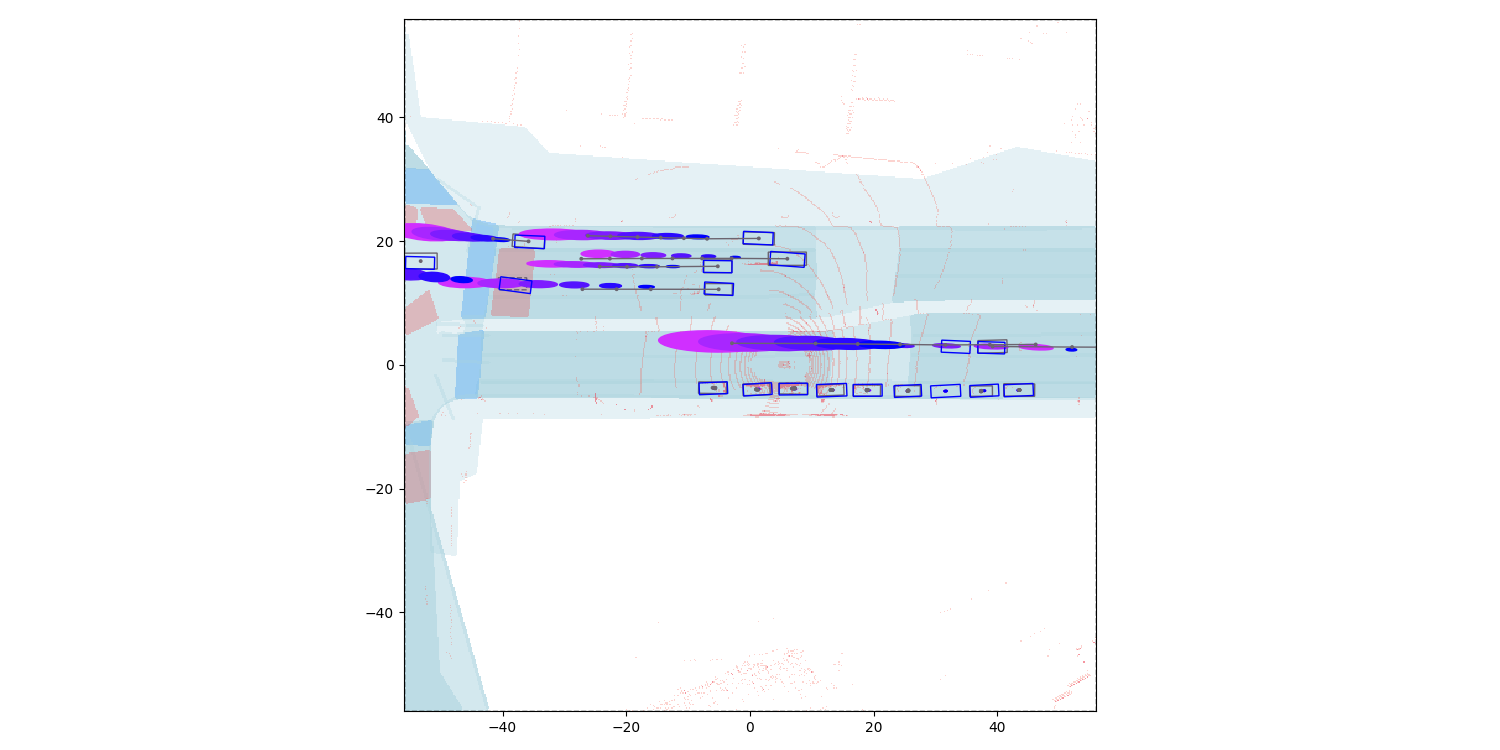}
\end{subfigure}%
\begin{subfigure}{.33\textwidth}
    \centering
    \includegraphics[width=0.97\textwidth, trim={12cm 3.0cm 12cm 3.0cm}, clip]{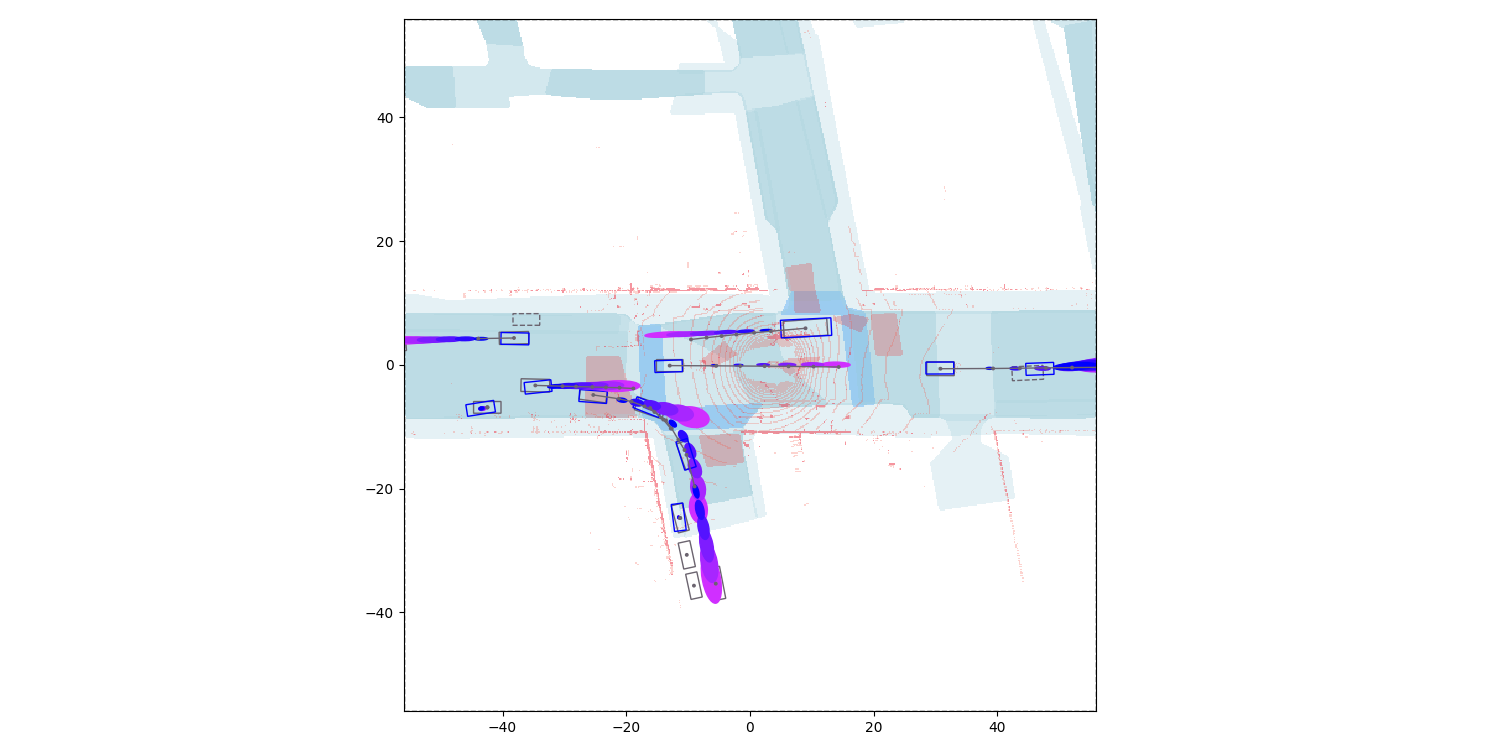}
\end{subfigure}%
\begin{subfigure}{.33\textwidth}
    \centering
    \includegraphics[width=0.97\textwidth, trim={12cm 3.0cm 12cm 3.0cm}, clip]{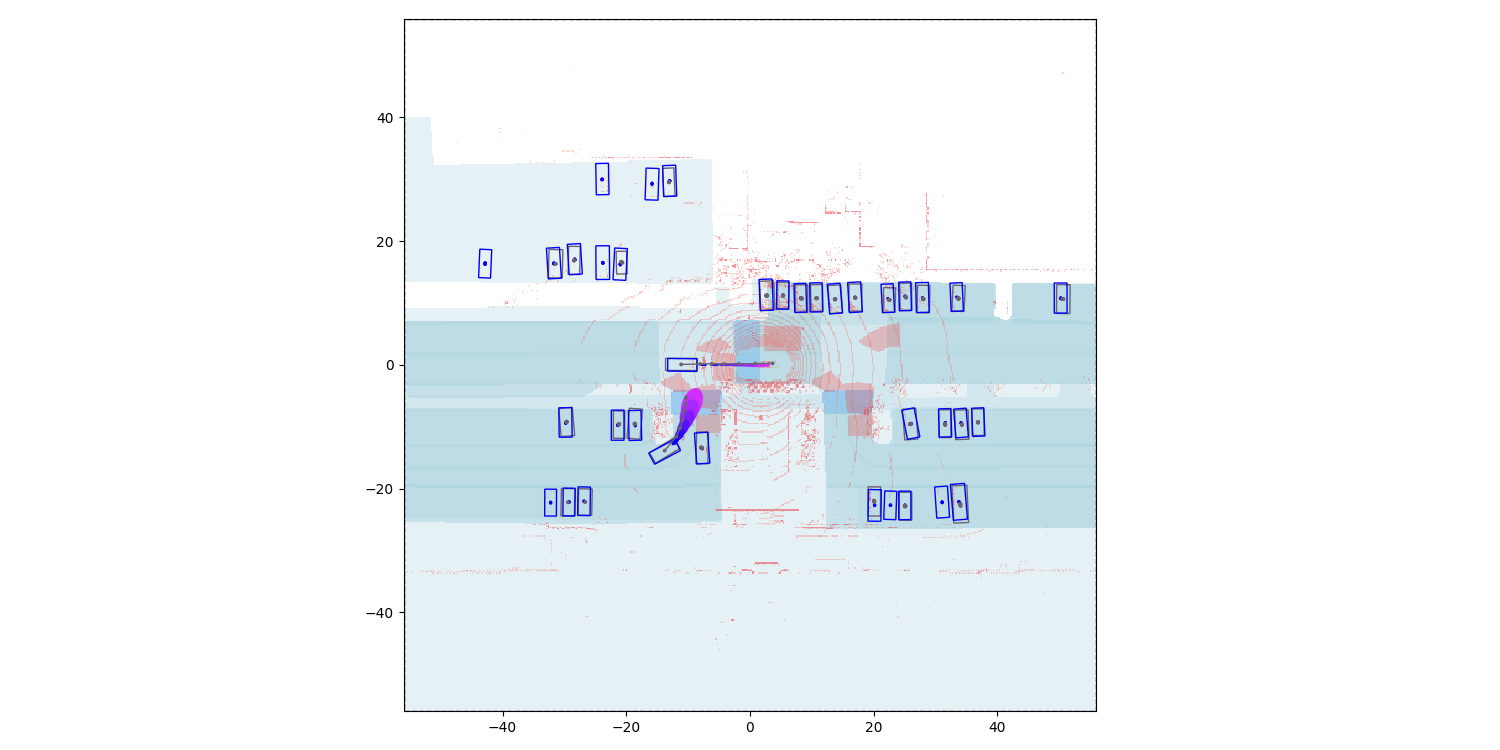}
\end{subfigure}
\vspace{0.2cm}
\begin{subfigure}{.33\textwidth}
    \centering
    \includegraphics[width=0.97\textwidth, trim={12cm 3.0cm 12cm 3.0cm}, clip]{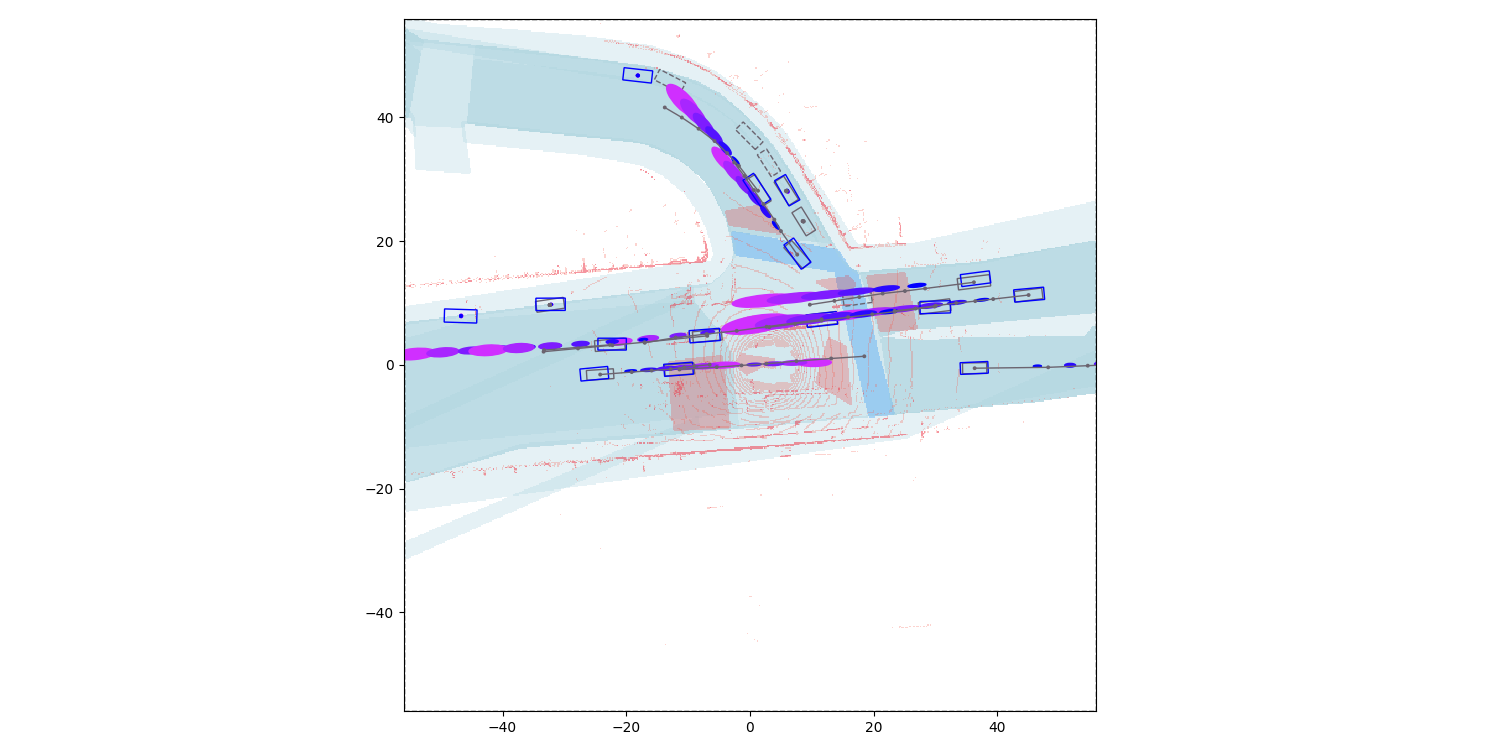}
\end{subfigure}%
\begin{subfigure}{.33\textwidth}
    \centering
    \includegraphics[width=0.97\textwidth, trim={12cm 3.0cm 12cm 3.0cm}, clip]{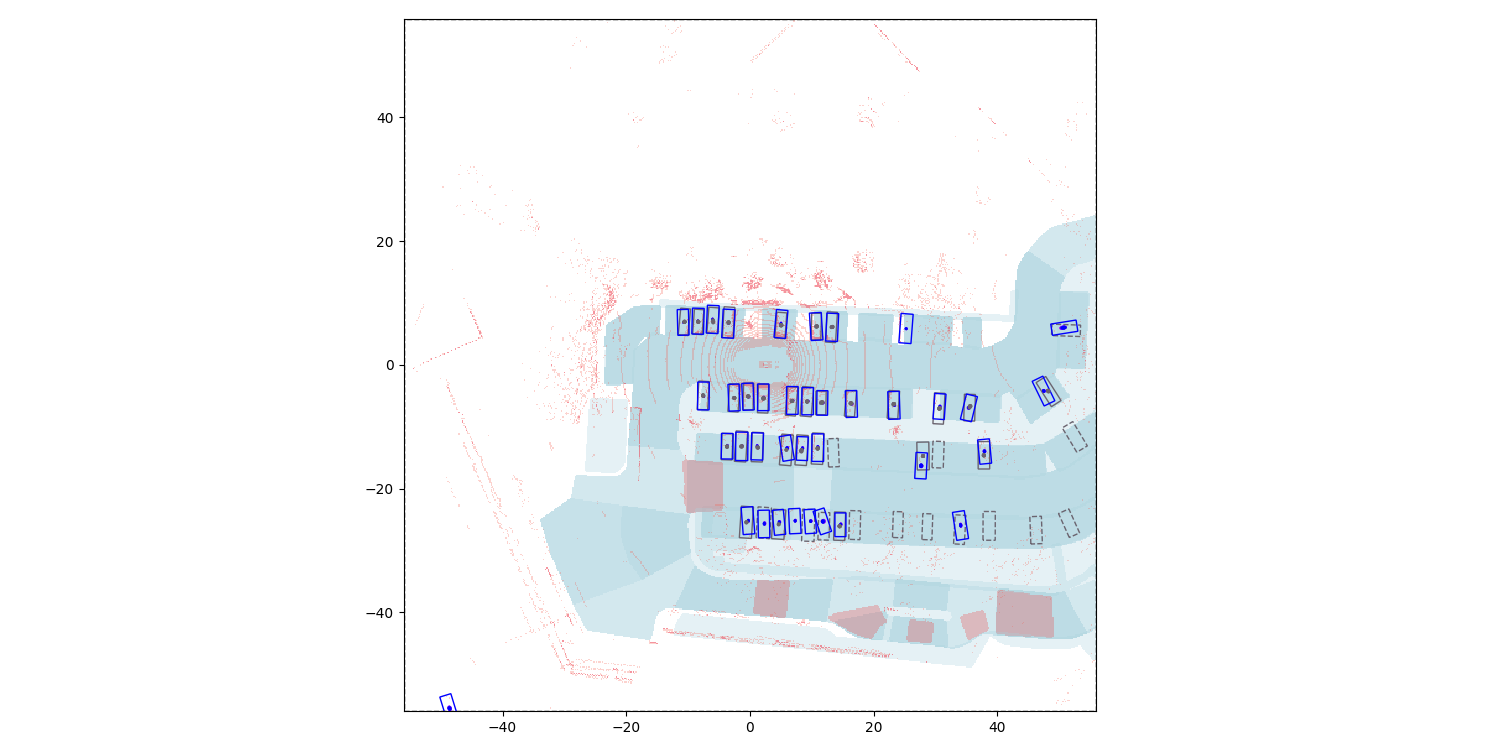}
\end{subfigure}%
\begin{subfigure}{.33\textwidth}
    \centering
    \includegraphics[width=0.97\textwidth, trim={12cm 3.0cm 12cm 3.0cm}, clip]{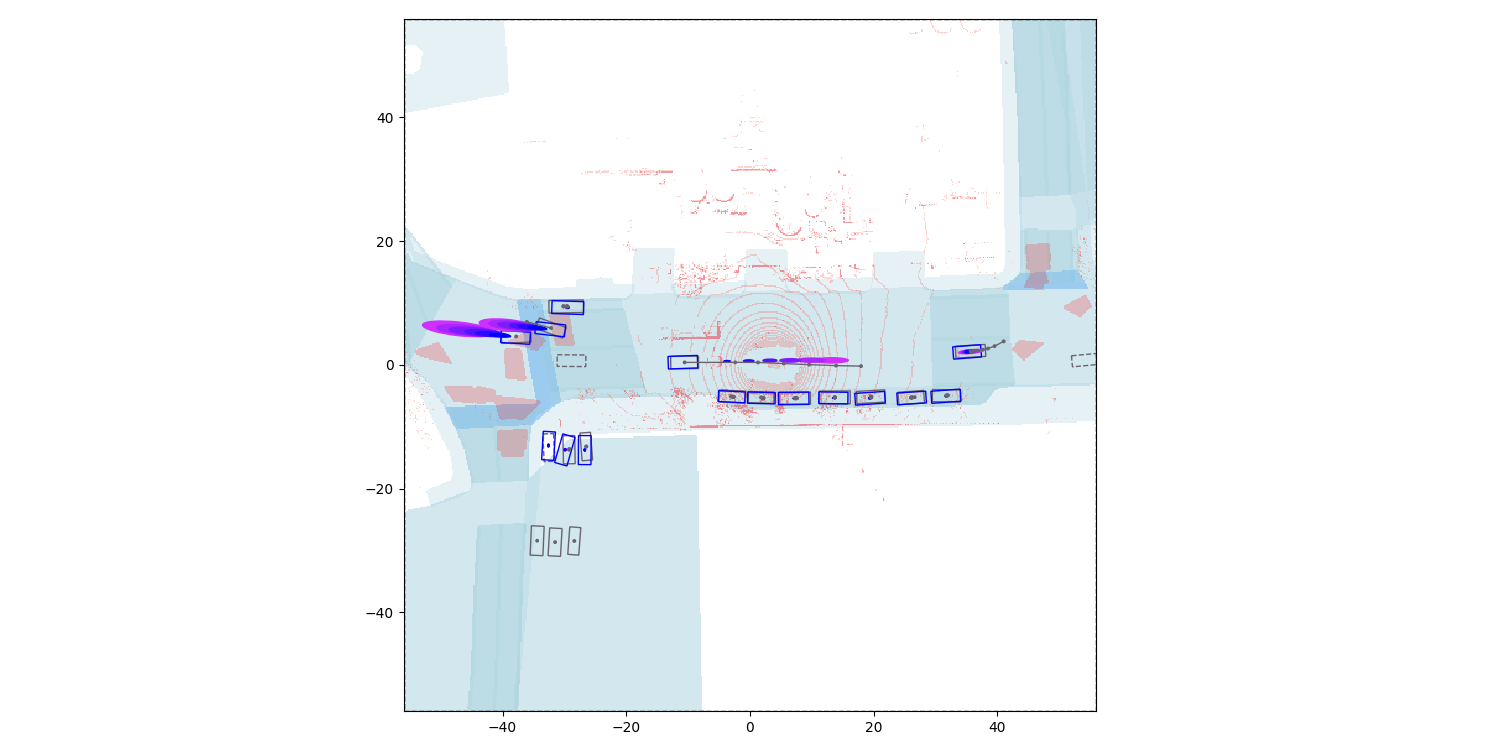}
\end{subfigure}
\vspace{0.2cm}
\begin{subfigure}{.33\textwidth}
    \centering
    \includegraphics[width=0.97\textwidth, trim={12cm 3.0cm 12cm 3.0cm}, clip]{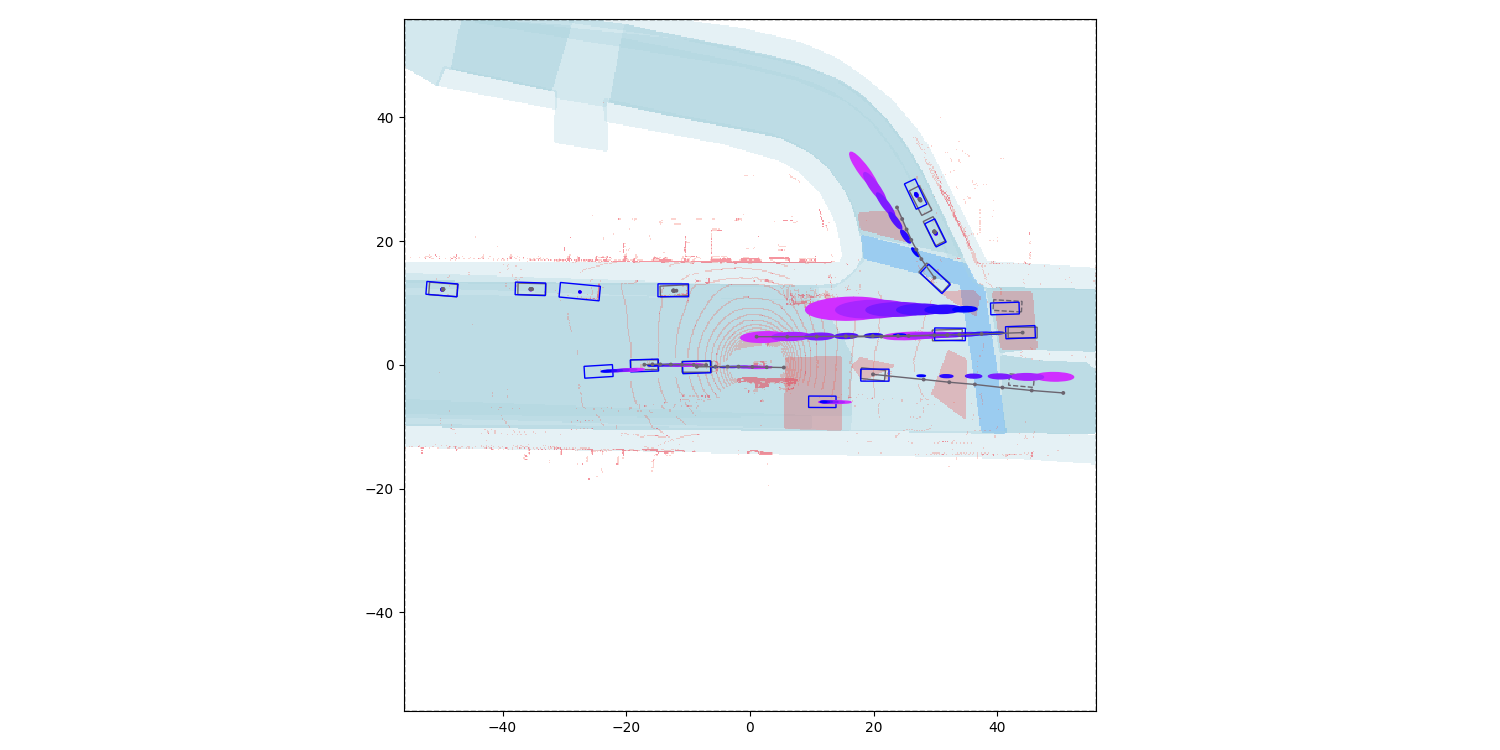}
\end{subfigure}%
\begin{subfigure}{.33\textwidth}
    \centering
    \includegraphics[width=0.97\textwidth, trim={12cm 3.0cm 12cm 3.0cm}, clip]{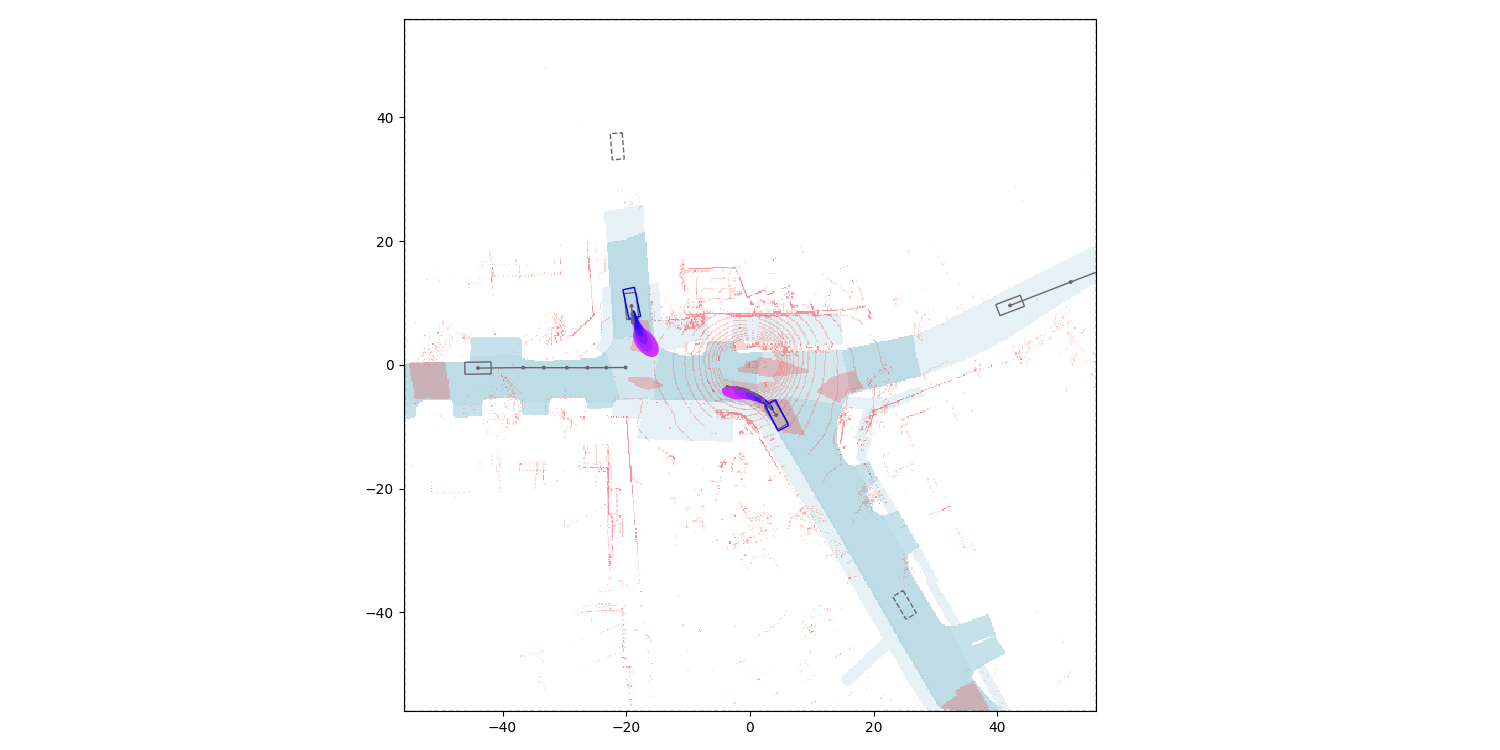}
\end{subfigure}%
\begin{subfigure}{.33\textwidth}
    \centering
    \includegraphics[width=0.97\textwidth, trim={12cm 3.0cm 12cm 3.0cm}, clip]{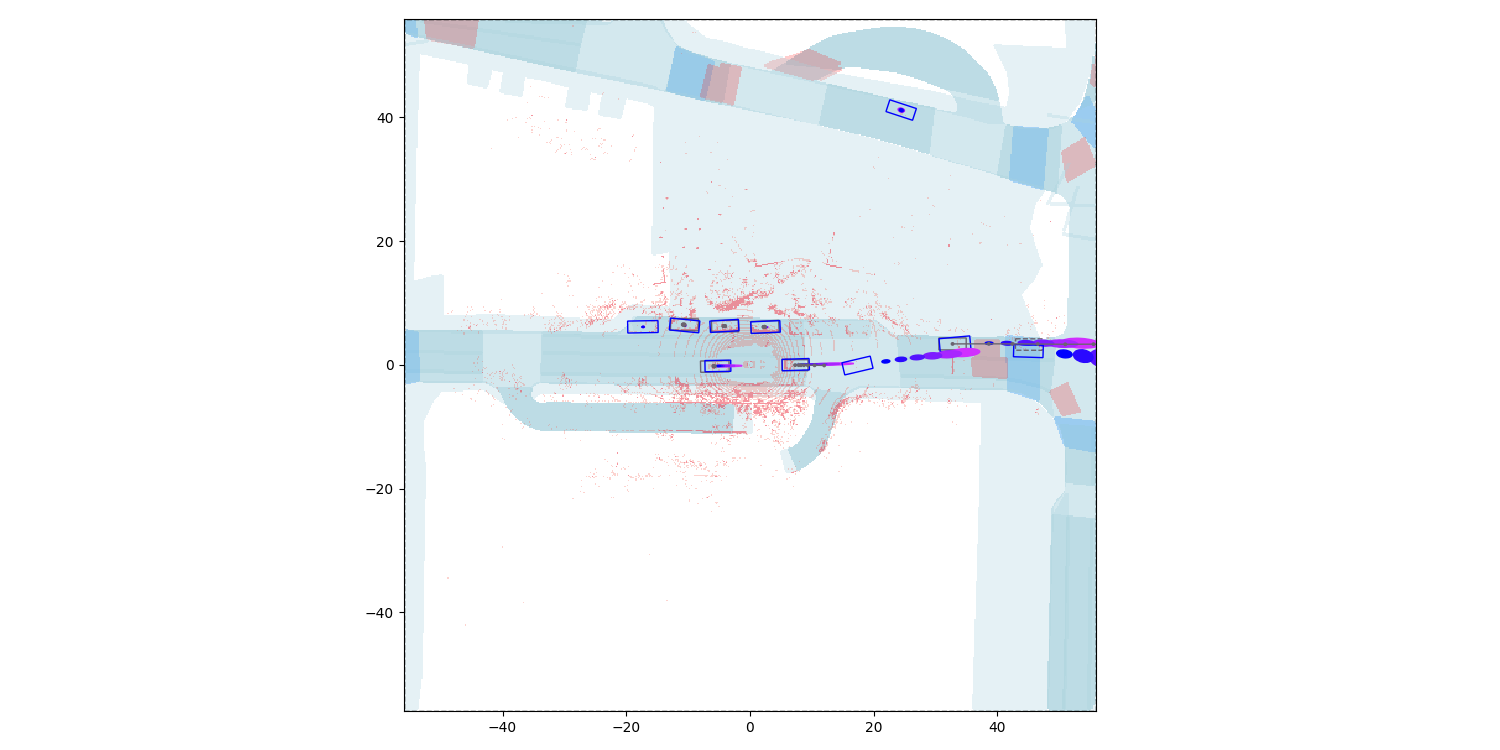}
\end{subfigure}
\caption[short]{\textbf{[\nuscenes{}] Additional qualitative results} of our \ourmodelshort{}. Detections are shown as blue bounding boxes. Probabilistic motion forecasts are shown as ellipsis (corresponding to one standard deviation of a bivariate gaussian) where variations in color indicate different future time horizons (from 0 seconds in blue to 3 seconds in pink).
Ground-truth boxes and future waypoints are displayed in gray. A dashed gray box means the object is occluded. Last row shows failure modes.}
\label{fig:additional_qualitative_nuscenes}
\end{figure*}

First, we present additional qualitative results in the \ourdataset{} dataset in Fig.~\ref{fig:additional_qualitative}. We can see how \ourmodelshort{} is able to predict very accurate detection and motion forecasts in a wide variety scenarios in terms of density of actors, lane-graph topologies, interactions and high-level actions (e.g. illegal u-turn or cars pulling out into non-mapped driveways). The last row of examples includes examples of our main failure mode: our model predicts a plausible trajectory in a multi-modal situation, but in the ground-truth another mode was executed. Finally, we present additional qualitative results in the \nuscenes{} dataset in Fig.~\ref{fig:additional_qualitative_nuscenes}, showing that our method can generalize to other datasets. However, we can appreciate in the last row (failure modes), that the detection is somewhat less reliable, mainly due to the 32-beam LiDAR sensor in contrast to the 64-beam one used in \ourdataset{}.

\end{document}